\documentclass{article}

\usepackage[preprint]{neurips_2025}

\usepackage[hidelinks]{hyperref}
\usepackage{url}
\usepackage{enumitem}
\usepackage[small]{caption}
\usepackage{subcaption}
\usepackage[compact]{titlesec}
\titlespacing{\section}{0pt}{1ex}{0.5ex}
\titlespacing{\subsection}{0pt}{0.5ex}{0ex}
\titlespacing{\subsubsection}{0pt}{0.5ex}{0ex}
\usepackage{setspace}

\usepackage{amsmath}
\allowdisplaybreaks
\usepackage{amssymb}
\usepackage{amsfonts}
\usepackage{bbm}

\usepackage[table]{xcolor}

\usepackage{booktabs}
\usepackage{tabularx}
\usepackage{multirow}
\usepackage{makecell}
\usepackage{arydshln}
\usepackage{colortbl}

\usepackage{graphicx}
\usepackage{float}
\usepackage{wrapfig}

\usepackage{pifont}
\usepackage{fontawesome5}
\usepackage{xspace}
\newcommand{\cmark}{\textcolor{green}{\ding{51}}}%
\newcommand{\xmark}{\textcolor{red}{\ding{55}}}%

\newcommand{\model}{\textsc{ZooWork-ShopRanker}\xspace}
\newcommand{\bench}{\textsc{ShopRank-Bench}\xspace}

\usepackage[noabbrev,capitalize]{cleveref}
\crefname{equation}{equation}{equations}
\crefname{section}{section}{sections}
\crefname{footnote}{footnote}{footnotes}
\crefname{line}{line}{lines}
\crefname{lstlisting}{listing}{listings}
\Crefname{lstlisting}{Listing}{Listings}

\AtBeginDocument{%
  \addtolength\abovedisplayskip{-0.25\baselineskip}%
  \addtolength\belowdisplayskip{-0.25\baselineskip}%
  \addtolength\abovedisplayshortskip{-0.25\baselineskip}%
  \addtolength\belowdisplayshortskip{-0.25\baselineskip}%
}

\newcommand{\codefont}{\fontfamily{lmtt}\selectfont}

\usepackage{listings}
\lstdefinestyle{productjson}{
    basicstyle={\codefont\small},
    xleftmargin={6pt},
    xrightmargin={6pt},
    frame=tb,
    tabsize=2,
    showtabs=false,
    showspaces=false,
    showstringspaces=false,
    extendedchars=true,
    breaklines=true,
    columns=fullflexible,
    keepspaces=true,
    captionpos=b,
    backgroundcolor=\color{green!5},
    aboveskip={0.8\baselineskip},
    belowskip={0.2\baselineskip},
    morestring=[b]",
    stringstyle=\color{black},
}

\lstdefinestyle{tracebox}{
    basicstyle={\codefont\footnotesize},
    xleftmargin={6pt},
    xrightmargin={6pt},
    frame=tb,
    tabsize=2,
    showtabs=false,
    showspaces=false,
    showstringspaces=false,
    extendedchars=true,
    breaklines=true,
    columns=fullflexible,
    keepspaces=true,
    captionpos=b,
    backgroundcolor=\color{blue!4},
    aboveskip={0.8\baselineskip},
    belowskip={0.2\baselineskip},
    morestring=[b]",
    stringstyle=\color{black},
}

\usepackage[textwidth=2.7cm]{todonotes}

\newcommand\myshade{85}
\colorlet{myurlcolor}{blue}
\hypersetup{
  citecolor  = black,
  urlcolor   = myurlcolor!\myshade!black,
  colorlinks = true,
}

\title{\model: An Open, Preference-Aligned E-Commerce Reranker}

\author{%
  Siqiao Xue, Shuxuan Liu, Ning Hu\\[0.6em]
  ZooWork Team\\[0.6em]
  \textbf{%
    \href{https://serendipityoneinc.github.io/look-bench-page/}%
      {\textcolor{blue!60!black}{\faGlobe\enspace{Project Page}}}%
    \quad
    \href{https://huggingface.co/collections/srpone/rerankers-in-e-commerce-69c4a9acb3eb3f8284d6c0c8}%
      {\textcolor{blue!60!black}{\faCogs\enspace{Models}}}%
    \quad
    \href{https://github.com/SerendipityOneInc/look-bench}%
      {\textcolor{blue!60!black}{\faGithub\enspace{Code}}}%
    \quad
    \href{https://huggingface.co/datasets/srpone/zoowork-shoprank-bench}%
      {\textcolor{blue!60!black}{%
        \raisebox{-0.22ex}{\includegraphics[height=1em]{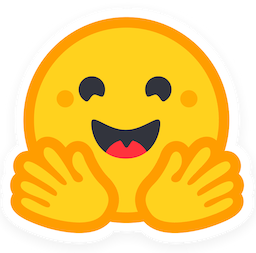}}%
        \enspace ShopRank-Bench}}%
  }%
}

\begin{document}

\maketitle

\begin{abstract}
Open rerankers trained for general web retrieval transfer imperfectly to
e-commerce, where ranking decisions depend not only on topical relevance but
also on user preferences, product constraints, and comparative product fit.
These preference signals are difficult to supervise at scale: real search
traffic provides authentic queries and candidates but no clean pairwise labels.
We present \model, a family
of e-commerce rerankers (0.6B, 4B, and 8B) aligned to judge-labeled shopping
preference.
Training pairs are labeled by a panel of reasoning large language models
(LLMs) from different families acting as a preference oracle, with
position-debiased judgments and agreement tiers, and the rerankers are trained
on these labels. The aligned
8B flagship then serves as a distillation teacher for the efficient 4B and 0.6B
models, which are fit to its scores and sharpened on judged pairs. To measure
progress, we introduce \bench, a contamination-limited benchmark of
${\sim}10{,}000$ private-traffic preference pairs in both text formats, tiered
by how many judge families committed to each label.
\model-8B and -4B significantly outperform the strongest open
reranker baseline, every model significantly beats its own un-aligned base, and
\model-0.6B beats its size peer; the gains hold in both formats
and extend to common MTEB benchmarks. We release the
models and the dual-format \bench to facilitate further research.
\end{abstract}

\section{Introduction}
\label{sec:intro}

Reranking is the final quality gate in an e-commerce search stack: retrieval
finds plausible products, and a reranker decides which of them best satisfies
the shopper. Open rerankers such as BGE-Reranker-v2-m3~\citep{chen2024bgem3},
Jina rerankers~\citep{sturua2024jinarerankerv3,jinaembeddingsv42025}, and the
Qwen3-Reranker family~\citep{qwen3embedding2025} perform strongly on public
retrieval benchmarks such as BEIR~\citep{thakur2021beir} and
MS~MARCO~\citep{nguyen2016msmarco}. Yet topical relevance is not user
preference. Between two individually relevant products, the better choice can
turn on product type, an explicit budget, comparative fit, and the shopper's
personal context.

\begin{wrapfigure}{r}{0.47\textwidth}
\vspace{-\intextsep}
\centering
\includegraphics[width=\linewidth]{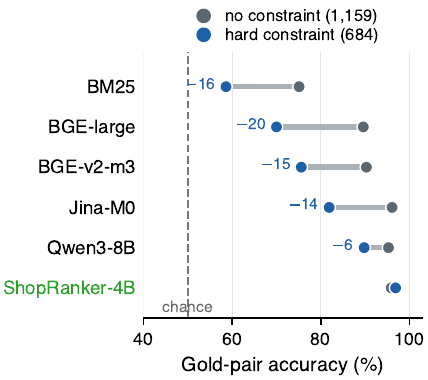}
\caption{Open rerankers score far lower on pairs whose query states a hard
constraint. Accuracy on the 1{,}843 gold-tier pairs
(\Cref{sec:benchmark_construction}), split by whether the query states an
explicit constraint; each segment is the accuracy on that subset. These are two
different sets of pairs, not the same
pairs with a constraint added and removed, so the difference is an association
rather than a measured effect of the constraint. The dashed line marks chance
(50\%), since every pair is a two-way choice. Baselines are 14--20 points lower,
the un-aligned base is more robust, and \model{} is flat. Per-subset numbers,
including the budget slice where BM25 falls below chance, are in
\Cref{tab:constraint_breakdown}.}
\label{fig:constraint_gap}
\end{wrapfigure}

What counts as the better product is not arbitrary. Following the judgment
protocol we adopt (\Cref{lst:judge_prompt}), a preference decision proceeds in
two tiers: a candidate must first satisfy the query's explicit \emph{hard
constraints} (the intended product type, a stated budget, the intended
recipient, and any explicit exclusions), and only among the survivors is the
softer preference evidence (style, color, fit, quality) weighed. This ordering
is \emph{query-conditional}: the hard constraints are whatever a given query
states, not a fixed global ranking of attributes. In fact, imposing a
hand-designed attribute-priority hierarchy is a poor optimization target that
anti-correlates with judged preference (\Cref{app:ahp}); what works is
learning this two-tier procedure from judge-labeled preference.
\Cref{tab:motivating} shows the failure mode this exposes in open rerankers: on
gold-tier pairs where a single hard constraint decides, strong open baselines
collapse to lexical or topical token matching and select the
constraint-violating product, whereas \model{} honors the constraint. This is
systematic rather than anecdotal: open baselines score 14--20 points lower on
the pairs whose query states a constraint than on those that state none, while
our models are flat
(\Cref{fig:constraint_gap}). The effect is sharpest for explicit budgets, where
the lexical baseline falls below chance and the cross-encoders lose $26$--$34$
points without reaching it (\Cref{tab:constraint_breakdown}), because a relevance
ranker has no reason to prefer the cheaper of two equally relevant products. Honoring an arbitrary
\emph{stated} budget, rather than this organic prefer-cheaper signal, remains a
harder and separate problem (\Cref{app:budget}).

\begin{table}[t]
\centering
\small
\setlength{\tabcolsep}{4pt}
\begin{tabularx}{\textwidth}{@{}
  >{\raggedright\arraybackslash\hsize=1.05\hsize}X
  >{\raggedright\arraybackslash\hsize=0.55\hsize}X
  >{\raggedright\arraybackslash\hsize=1.40\hsize}X
  @{}}
\toprule
{\sc Query} & {\sc Decisive constraint} & {\sc Candidate products} \\
\midrule
\emph{``shape sorter''} & product type
& \cmark~Fisher-Price shape-sorter \textbf{toy} \newline
  \xmark~office \textbf{mail sorter} \\
\addlinespace[4pt]
\emph{``jumpsuit under \$100 boho''} & explicit budget
& \cmark~\textbf{\$44} jumpsuit \newline
  \xmark~\textbf{\$239} ``bohemian'' one \\
\addlinespace[4pt]
\emph{``jacket for dad vintage''} & intended recipient
& \cmark~\textbf{men's} jacket \newline
  \xmark~\textbf{women's} ``vintage'' jacket \\
\addlinespace[4pt]
\emph{``swim trunks without mesh lining''} & explicit exclusion
& \cmark~\textbf{non-mesh} trunk \newline
  \xmark~\textbf{mesh}-lined trunk \\
\bottomrule
\end{tabularx}
\caption{Open rerankers rank by topical relevance, not shopper
preference. Four gold-tier \bench{} pairs in which one explicit constraint
decides the winner (\cmark{} preferred, \xmark{} rejected). All four open
baselines---BM25, BGE-Reranker-large, BGE-Reranker-v2-m3, Jina-m0---pick the
violating product on every pair, each lured by a surface token match;
\model-8B and -4B honor the constraint on all four, \model-0.6B on two.}
\label{tab:motivating}
\end{table}

This preference gap is hard to supervise at scale: real traffic offers
authentic queries and candidates but no clean pairwise labels. We obtain them
from reasoning-capable large language model (LLM) judges used as a preference
oracle---cross-checked across model families, scored in both presentation
orders, and separated by agreement strength---yielding pairs suited to direct
pairwise preference training. We also tried concentrating labeling at the
current decision boundary by mining on-policy, and report the ceiling it ran
into: the frontier becomes judge-ambiguous rather than merely hard to mine
(\cref{sec:on_policy}).

Evaluation must also reflect deployment, where public sets risk pretraining
contamination, easy pairs hide model differences, and a single serialization
rewards formatting artifacts. We therefore build \bench from the private search
traffic of Gensmo (\url{https://studio.gensmo.com/}), a ZooWork commercial
search engine indexing billions of shop products, whose retrieval stack has
already supported open benchmarks and models for fashion
search~\citep{gao2026lookbench,xue2026zooclaw}. We release \bench in both
structured and natural-language formats, alongside attribute-hierarchy and
budget diagnostics, and report every preference-track result with intervals
and paired significance tests that account for the many pairs sharing a query;
the diagnostic and MTEB tables report point estimates.

We present \model, a family of open e-commerce rerankers (0.6B, 4B, and 8B)
aligned to shopping preference. On the
contamination-limited \bench, our 8B and 4B significantly beat the
strongest open reranker baseline and the 0.6B beats its size peer, in both formats
and on common MTEB tasks~\citep{muennighoff2022mteb}. We report serving cost
alongside accuracy---alignment is free at inference, and on structured product
text the distilled 0.6B matches a 4B base at $2.8\times$ the throughput (on
natural-language text the 4B base keeps a significant edge)---and further
analyses show the
gains hold in both product-text formats, that explicit query intent overrides a
conflicting user profile, and isolate where alignment does and does not install
constraint-following.

We release all three model sizes, the dual-format \bench, and code to
facilitate future research.
An optimized commercial version is available as a ZooWork API at
\url{https://zoodata.ai/en/api-docs}.

\section{Related Work}
\label{sec:related_work}

\paragraph{Open rerankers.}
The models named above divide by architecture, and the split matters for more
than taxonomy. Encoder-only cross-encoders score a pair with a dedicated
classification head, as in BGE-Reranker-v2-m3~\citep{chen2024bgem3} and the
Jina rerankers~\citep{sturua2024jinarerankerv3,jinaembeddingsv42025}. Decoder
rerankers instead read out designated yes/no or score tokens from a language
model, as in Qwen3-Reranker~\citep{qwen3embedding2025,yang2025qwen3}, the family
we build on; \Cref{app:latency} quantifies what carrying a full language model
to emit one scalar costs at serving time. Multimodal systems such as Jina-m0 and
Qwen vision-language models~\citep{Qwen2VL,Qwen-VL} extend the setup to product
images.

\paragraph{Scoring without generating.}
A reranker needs a decision, not prose, so the score can be read from the logits of
a fixed answer vocabulary in one forward pass. Qwen3-Reranker does this with
\texttt{yes}/\texttt{no} tokens~\citep{qwen3embedding2025}; setwise rankers extend
it to several candidates in a shared context and compare pointwise, pairwise,
listwise and setwise prompting on the efficiency--effectiveness
trade-off~\citep{zhuang2024setwise}; instruction distillation compresses the
expensive variants into cheaper students~\citep{sun2023instructiondistill}. The
same pattern has been productised as parallel constrained
decoding~\citep{typesafe2026jev}. We adopt it rather than propose it: every score
in this paper is produced this way, which is what makes evaluating a 27B model on
$10{,}511$ pairs affordable, and it lets us report the LLM reference and the
rerankers on one cost axis (\Cref{app:llm_reference}).

\paragraph{Preference optimization and LLM judges.}
DPO~\citep{rafailov2023dpo} converts pairwise preferences into a stable
classification objective. Its connection to ranking losses is especially
direct for discriminative rankers~\citep{jin2025larpo}: the policy output is a
scalar relevance score rather than generated text. We use multiple
reasoning-capable LLM families as scalable annotators, require agreement, and
judge both candidate orders to reduce position effects. This cross-family
protocol also separates the training panel from an additional evaluation
family, reducing circularity.

\paragraph{Domain-specific benchmarks.}
High-quality evaluation data is a recurring bottleneck for domain-specific
models. Recent efforts pair curated benchmarks with dedicated models in
domains such as fashion retrieval~\citep{gao2026lookbench,xue2026zooclaw} and code
search~\citep{xue2026coreb}. \bench shares their motivation:
carefully constructed domain evaluation exposes limitations that
general-purpose benchmarks cannot.

\section{The \texorpdfstring{\bench}{ShopRank-Bench} Benchmark}
\label{sec:data}

\bench is the centerpiece of our evaluation. It contains ${\sim}10{,}000$ hard
preference pairs derived from Gensmo's private search traffic. The pairings and
their preference labels have never been public, so unlike ESCI-style public
sets~\citep{reddy2022esci} the answers cannot have been memorised in
pretraining; we use \emph{contamination-limited} in that sense throughout. Single
queries, and the public catalogue attributes the product text is rendered from,
may well appear elsewhere on the web---what is unavailable is which candidate a
judge panel preferred. \Cref{app:contamination} reports the
overlap audit against our own training slice and checks that no trivial
surface heuristic solves the benchmark; \Cref{app:privacy} describes the
released fields and confirms they carry no personal information.
\bench is organized as a suite (\Cref{tab:bench_overview}):
\begin{itemize}[leftmargin=*,itemsep=1pt,topsep=2pt]
  \item a \emph{preference} track of judge-labeled pairs, the headline
  metric, released in both structured and natural-language product formats
  (\Cref{sec:dual_format}); and
  \item two \emph{diagnostic} tracks, attribute hierarchy (AHP) and explicit
  budget (budget), that probe constraint-following behavior
  (\Cref{sec:diag_construction,sec:ahp}).
\end{itemize}
The diagnostic tracks are scored separately and never folded into the
preference metric, because their controlled labels test rule-following rather
than judged preference. One illustrative record per track is given in
\Cref{app:track_examples}. The training corpus, which shares this substrate
but none of the benchmark's queries or pairs, is described with the training
recipe in \Cref{sec:train_setup}.

\begin{table}[t]
\centering
\small
\begin{sc}
\begin{tabularx}{\textwidth}{@{}lrcc>{\raggedright\arraybackslash}X>{\raggedright\arraybackslash}X@{}}
\toprule
 & & \multicolumn{2}{c}{Avg.\ tokens} & & \\
\cmidrule(lr){3-4}
Track & Pairs & query & doc & Label source & Purpose \\
\midrule
Preference & 10,511 & 6.7 & 99 & Cross-family LLM judges, no conflict at any tier: 1,843 gold (3/3 committed), 4,445 silver (2/3), 4,223 bronze (1/3) & Headline metric; judged preference \\
AHP & 1,500 & 7.4 & 93 & Hand-designed attribute-priority hierarchy (controlled construction) & Diagnostic; hierarchy-following \\
Budget & 1,302 & 10.4 & 100 & Programmatic $\mathrm{price}\le\mathrm{budget}$ oracle & Diagnostic; constraint-following \\
\bottomrule
\end{tabularx}
\end{sc}
\caption{Overview of the three \bench tracks; lengths are mean query and
product-text token counts. The preference track is the headline metric
(released in structured and natural-language formats); the attribute-hierarchy
(AHP) and explicit-budget tracks are diagnostics, scored separately.}
\label{tab:bench_overview}
\end{table}

\subsection{Shared substrate: traffic, retrieval, and product text}
\label{sec:source_traffic}

All three tracks are built on the same substrate: real user search sessions
logged by Gensmo's production engine. Queries in this traffic are natural and
often messy (e.g., ``sonic socks for boys 4--6 years'', ``printer for my moms
house she needs something simple not too big''), in contrast to the templated
or cleaned queries of public e-commerce sets. The traffic spans general
e-commerce (apparel, electronics, beauty, baby and kids, home, and
consumables) rather than a single vertical, matching the deployment scope of
the reranker, though it leans toward apparel (category distribution in
\Cref{app:categories}).

For each query, candidates are the top-ranked products returned by the
production retrieval stack, so every pair compares products a deployed system
actually surfaced. Each product record carries structured catalog attributes
(product type, color, audience, style, brand, material, occasion, price,
and so on), which we render into a canonical pipe-delimited product text; examples are in \Cref{app:schema}. This
retrieval and rendering pipeline is identical across tracks.

The tracks diverge only at the query and label stages. The \emph{preference}
track keeps traffic queries verbatim, pairs candidates by their production
ranks, and obtains labels from an LLM judge panel
(\Cref{sec:benchmark_construction}). The \emph{diagnostic}
tracks instead control both: queries are LLM-rewritten, intent-conditioned
variants of traffic queries, and labels come from rules rather than judges
(a hand-designed attribute hierarchy for AHP, a programmatic budget oracle
for the budget track; \Cref{sec:diag_construction}).

\paragraph{Query profile.}
Preference-track queries are drawn from a pool of roughly 4{,}700 unique
traffic strings, of which 2{,}991 carry at least one decisive pair into the
released benchmark (\Cref{sec:benchmark_construction}); they are short and underspecified (mean 6.7 tokens, up to
28) and span the category breadth described above. AHP queries are 1,495
intent-conditioned variants, and budget queries are 398 budget-explicit
variants each stating a numeric price cap. Per-level pair counts and one
representative query per track are given in \Cref{app:track_examples}.

\subsection{Preference track: construction, judging, and release}
\label{sec:benchmark_construction}

\paragraph{Pairing.}
We form difficult comparisons from candidates adjacent in the existing
production ranking: adjacent candidates are the ones a deployed system already
treats as near-equivalent, so separating them is exactly where a reranker must
add value, whereas randomly sampled pairs are dominated by easy contrasts that
hide model differences.

\paragraph{Judging protocol.}
Each pair is independently assessed by three model families:
Qwen3.5-122B, Gemma-4-31B, and DeepSeek-V4-Pro. Each judge sees both candidate
orders to reduce position bias and must follow a constraint-first protocol:
state the query's hard constraints, check each product against every
constraint, weigh the remaining preference evidence, and only then decide or
declare a tie (the full instruction is in \Cref{app:judge_protocol}).
Inter-family agreement, rather than confidence reported by a single judge, is
our label-credibility metric. DeepSeek-V4-Pro contributes only to benchmark
labels and never to training labels (\Cref{sec:train_setup}), so the benchmark
is not judged exclusively by the families that supervised the models.

\paragraph{Filtering and agreement tiers.}
A judge returns one of three verdicts per pair: it picks a product, or it
abstains by declaring a tie. A judge counts as committing only when both
presentation orders agree on the same product; one that picks different
products in the two orders is counted as abstaining. We keep a pair when the judges that committed all
chose the \emph{same} product and none contradicts them; abstentions are
permitted. Of 23,000 judged pairs, 10,511 survive this rule (45.7\%); the
remaining 12,489 are discarded: 12,350 are unanimous ties, and 139 are
outright conflicts in which committed judges named different products. The high discard
rate is itself evidence that adjacent-rank pairs sit at genuine decision
boundaries rather than being resolvable by surface relevance.

Because abstention is permitted, ``no conflict'' spans labels of very different
strength, and we tier the release by \emph{how many} families actually
committed:
\begin{itemize}[leftmargin=*,itemsep=1pt,topsep=2pt]
  \item \emph{gold} (1,843 pairs): all three families committed and agreed;
  \item \emph{silver} (4,445): two committed and agreed, one abstained;
  \item \emph{bronze} (4,223): one committed and two abstained, so the label
  rests on a single family.
\end{itemize}
Reporting these separately matters: the bronze tier is 40\% of the benchmark
and every system is 20--26 points weaker on it than on gold
(\Cref{tab:main_structured}), so an aggregate score is dominated by the pairs
carrying the weakest labels. The per-judge verdicts are released with each
record, so this tiering can be recomputed by anyone.

The panel is also not symmetric in how often each family commits: of the
10,511 released pairs, Gemma-4-31B commits on 92.7\%, DeepSeek-V4-Pro on
66.1\%, and Qwen3.5-122B on only 18.6\%, abstaining on the rest. Over all
23,000 judged candidates the same rates are 43.0\%, 30.8\% and 8.5\%. The bronze tier is therefore largely
Gemma-decided. We report it as a distinct tier rather than dropping it, and
\Cref{app:circularity} checks how much of our margin over the baselines
survives on pairs a training-disjoint family also decided.

\paragraph{Dual-format release.}
\phantomsection\label{sec:dual_format}
Every preference-track candidate is released twice: once in the canonical
structured attribute schema and once as natural-language product text. The natural language is
model-rendered from the same attributes, not scraped prose; renderings vary
sentence order and attribute inclusion so a model cannot memorize a single
template. This paired design holds the query, the products and the label fixed,
but it does not isolate serialization alone: because attribute inclusion varies,
a prose view can omit attributes the structured view states---the rendering in
\Cref{lst:dual_format} drops audience, fit, occasion and season---and any of
those can be preference-determining. A format gap measured this way therefore
bounds sensitivity to rendering \emph{and} attribute omission together, and we
do not audit whether every rendering preserves the attributes its label turned
on. \Cref{lst:dual_format} shows the
same product in both formats.

\begin{lstlisting}[float=t, style=productjson, caption={Dual-format example: the same
product in the structured attribute schema (top) and one model-rendered
natural-language view (bottom). Both formats carry the same preference
label.},label=lst:dual_format]
Structured:
  style: casual | color: caviar | material: fleece |
  audience: women | product type: hoodie |
  brand: Reef | pattern: solid | fit: relaxed fit |
  length: regular | sleeve: long | occasion:
  everyday, school | season: fall | price: 34.99 |
  product title: Reef hoodie

Natural language:
  Reef hoodie. A caviar casual hoodie from Reef,
  around $34.99, made of fleece.
\end{lstlisting}

\subsection{Diagnostic tracks: construction and rationale}
\label{sec:diag_construction}

Both diagnostic tracks reuse the shared substrate of
\Cref{sec:source_traffic} (catalog products, production retrieval, canonical
product text) and depart from the preference track at the two stages noted
there: queries are intent-conditioned variants rather than verbatim traffic,
and labels come from explicit rules rather than the judge panel.

\paragraph{AHP track.}
Each of the 1,500 controlled pairs differs on exactly two attributes of the
hierarchy, and the
product satisfying the higher-priority attribute is labeled the winner. Priority
is \emph{intent-conditioned}: an LLM tags each query with a shopping intent
(navigational, price-driven, occasion, style, persona, or a default), and each
intent maps to a hand-designed attribute-priority order---a price-driven query
ranks \texttt{price} first, a navigational query ranks \texttt{brand} first,
and the default order is audience $>$ product type $>$ price $>$ style $>$
color. Pairs are mined from real retrieval: candidates are fetched from the same
private search index, and only pairs whose attribute difference is exactly two
are kept, so each pair isolates a single hierarchy decision under its intent. The rationale is
control: organic preference pairs entangle many attributes at once, whereas
this construction asks one question per pair: does the model rank the intent's
higher-priority attribute above the lower one? Because the labels encode a
hand-authored heuristic rather than observed preference, the track is
diagnostic only; indeed, \Cref{app:ahp} shows the heuristic anti-correlates
with judged preference. \Cref{lst:ahp_example} (\Cref{app:track_examples})
shows a record.

\paragraph{Budget track.}
The 1,302 budget pairs are drawn from the budget-explicit subset of the same
mined query pool (e.g., ``under \$50''): the budget is parsed from the query,
prices from the product text, and the label is the programmatic oracle
\(\mathrm{price}\le\mathrm{budget}\), giving clean supervision that requires
no judge. The track has two query-disjoint slices designed so that no
price-monotone shortcut can win both: a \emph{threshold} slice (853 pairs; one
product meets the stated budget, the other exceeds it) and a \emph{control}
slice (449 pairs; the budget is raised above both prices, so a secondary
attribute decides and the pricier product wins by construction). A pick-cheaper
policy scores 1.000/0.000 on threshold/control and pick-expensive the reverse,
so beating both slices requires treating the budget as a threshold. The
rationale is to separate soft price preference, which is judge-ambiguous and
subject to the clean-label ceiling of \Cref{sec:on_policy}, from explicit
budget compliance, which is programmatically decidable; this tests whether the
universal price failure in \Cref{tab:ahp_results} reflects missing capacity or
missing supervision (\Cref{app:ahp}). \Cref{lst:budget_example}
(\Cref{app:track_examples}) shows a threshold-slice record.

\section{Training \texorpdfstring{\model}{ZooWork-ShopRanker}}
\label{sec:models_training}

The released models are built in two different ways, and it matters which is
which. The flagship \model-8B is aligned directly: the Qwen3-Reranker-8B base is
trained on judge-labeled pairs with the pairwise preference loss of
\Cref{sec:dpo}, then given a final pass on mixed structured and
natural-language product text, a stage that a controlled comparison shows to be
neutral (\Cref{app:format_training}).
The two smaller models are not aligned
from their own bases at all. Instead the aligned 8B scores a large pool of
query--document pairs, the 0.6B and 4B are fit to those soft scores with binary
cross-entropy, and each is then sharpened on judged pairs with the same
pairwise objective (\Cref{sec:distillation}). So the 8B is the teacher
and the 4B is a student alongside the 0.6B, not a scaled-down copy of the 8B
recipe.

Two further stages appear in this section because they shaped the recipe, but do
not appear in the released small models: an on-policy refinement round
(\Cref{sec:on_policy}), which we report for the ceiling it exposed rather than
for its contribution, and a judged-pair alignment pass on the 4B, which the distillation
route superseded. \Cref{tab:ablation} decomposes that earlier
alignment-from-base pipeline, and each stage is reported alongside the negative
result that motivated it.

\subsection{Training setup}
\label{sec:train_setup}

\paragraph{Base models.}
\model-0.6B, -4B, and -8B are LoRA adapters~\citep{hu2022lora} on
Qwen3-Reranker-0.6B, -4B, and -8B~\citep{qwen3embedding2025}. Each
decoder-style reranker maps the official query--document prompt to \texttt{yes}
and \texttt{no} token logits whose difference is the scalar relevance score
(\Cref{lst:qwen_prompt}). No text is generated: the decision is read from the
logits of a fixed answer vocabulary in a single forward pass, the
\emph{parallel constrained decoding} pattern that scores candidate outputs
instead of decoding them token by token~\citep{zhuang2024setwise,typesafe2026jev}.
We use it for every score in this paper---the released rerankers, the open
baselines, and the LLM reference of \Cref{app:llm_reference}---which is what makes
benchmark-scale LLM evaluation affordable at all. We use this family because one architecture spans all
three scales---so one architecture spans teacher and students, and the aligned
8B can teach both smaller models---the bases already match or beat dedicated open
cross-encoders zero-shot, the binary-token score is exactly the scalar the
pairwise loss acts on, and the open weights let us release the adapters.

\paragraph{LoRA-only training.}
All training stages, from the initial judged-pair alignment through on-policy
refinement, distillation, and format mixing, update LoRA adapters only; the
base weights stay frozen throughout, and no stage uses full fine-tuning. This
is a deliberate choice, not a compute concession: a controlled comparison on
identical data and recipe shows LoRA matches or edges full fine-tuning at
roughly 1\% trainable parameters (\Cref{sec:param_efficiency}).

\begin{lstlisting}[float=t, style=tracebox, caption={Zero-shot evaluation prompt for the Qwen3-Reranker baselines. The yes/no logit difference is the relevance score.}, label=lst:qwen_prompt]
System: You are a product search relevance scorer. Given a query and a product,
        output a single token "yes" if the product is relevant, otherwise "no".
User:   Query: {query}
        Product: {structured_product_text}
        Relevant?
\end{lstlisting}

\paragraph{Training corpus.}
The training data comes from the same substrate as \bench
(\Cref{sec:source_traffic}) but from a disjoint slice of traffic; the
contamination audit of \Cref{app:contamination} confirms zero query overlap and
zero exact-pair overlap between this slice and the benchmark. Its core is approximately
4.4k judge-labeled preference pairs mixed with general retrieval replay data,
which preserves broad relevance behavior during preference alignment. Two
further components extend the corpus: on-policy refinement adds pairs the current
model scores as nearly equal (\Cref{sec:on_policy}), which the released models
do not use, and distillation scores
134,601 query--document examples with the aligned \model-8B as soft
labels for the 0.6B and 4B students (\Cref{sec:distillation}). All training text
uses the canonical structured product text of \Cref{app:schema}; the
mixed-format stage additionally mixes in natural-language renderings produced
by the same pipeline as the benchmark's NL view (\Cref{app:format_training}).
The corpus is proprietary and not part of the release; \bench is the released
evaluation artifact.

\paragraph{Training judge.}
Every training label---the core pairs and those added on-policy---is set by a
two-family LLM judge panel, Qwen3.5-122B and Gemma-4-31B, under the same
constraint-first, both-orders protocol used to build \bench
(\Cref{app:judge_protocol}); a pair is retained only when both families agree
in both presentation orders. The third benchmark family, DeepSeek-V4-Pro, is
withheld from training, so the final \bench evaluation (\Cref{sec:experiments})
is not adjudicated by the models that supervised the pipeline.

\paragraph{Development evaluation.}
To monitor training, we hold out a development set drawn from the training substrate but
disjoint from both the training pairs and \bench. It has three slices,
\emph{easy} (5,076 pairs), \emph{hard-structured} (1,910), and \emph{hard-NL}
(1,910), where hard pairs are adjacent in production rank and hard-NL is their
natural-language rendering. Unless stated otherwise, every accuracy figure in
this section is measured on these development slices, which carry two-judge
labels; they track relative progress across stages but are not directly
comparable to the three-judge \bench numbers of \Cref{sec:experiments}.

Training hyperparameters are given in \Cref{app:hparams}.

\subsection{Training components}
\label{sec:pipeline}

We describe the components in the order they were developed. They are not a
single linear pipeline: which of them a given released model uses is set out
above.

\paragraph{Judge-labeled preference training.}
\label{sec:dpo}
For query \(q\), preferred product \(d^+\), rejected product \(d^-\), and
reranker score \(f_\theta\), we optimize the pairwise logistic loss
\begin{equation}
\label{eq:dpo_reranker}
\mathcal{L}
 =-\log\sigma\!\left(\beta\left[f_\theta(q,d^+)-f_\theta(q,d^-)\right]\right).
\end{equation}
This is the classical Bradley--Terry pairwise
objective~\citep{bradley1952rank}, equivalently RankNet~\citep{burges2005ranknet}
with a temperature, and the same loss used to fit reward models in RLHF. We
claim no novelty in it; what is ours is the data it is trained on
(\Cref{sec:data,sec:train_setup}). The training pairs are the judge-labeled
corpus of \Cref{sec:train_setup}, mixed with general replay.

\emph{Relation to DPO.}
Readers coming from preference optimization may ask where the reference policy
went. In generative DPO~\citep{rafailov2023dpo} the reference serves two
purposes: it parameterizes the implicit reward as a log-probability ratio, and
it regularizes the policy against degenerate text. Neither applies to a
discriminative reranker. The model emits the scalar reward directly (the yes/no
logit gap), so the reward reparameterization collapses to the score itself and
DPO reduces to \Cref{eq:dpo_reranker}---which is why no reference model appears
here. With no generated text to protect from collapse, the anchoring role falls
instead to LoRA's low-rank update and the general-relevance replay mixed into
training (\Cref{sec:train_setup}); dropping the reference is likewise effective
for generative models~\citep{meng2024simpo}.

\paragraph{On-policy refinement, and the ceiling it exposed.}
\label{sec:on_policy}
After initial judged-pair alignment, we mine the pairs the model finds hardest, those with the
smallest score margin \(\lvert s^+-s^-\rvert\), and send them to the training
judge panel of \Cref{sec:train_setup}. A first on-policy round gives a small
gain (development easy-slice accuracy 0.836 to 0.844), but further rounds stall
for a specific reason. Alignment polarizes the model's scores toward the
extremes, so the smallest-margin pairs are no longer genuinely borderline: they
are two-good or two-bad products that the judges themselves tie on. The supply
of clean, agreed labels therefore dries up (the yield of gold pairs from a
mining batch falls from 26\% to 6\%) even though the pairs are, by score,
maximally confusable. The frontier has become judge-ambiguous rather than merely
hard to mine: on-policy preference learning is clean-label-limited.

\paragraph{Distillation into the efficient 0.6B and 4B.}
\label{sec:distillation}
We score approximately 135k query--document examples with the aligned \model-8B
and fit each student to its soft scores using binary cross-entropy, mixing
structured and natural-language renderings of the same pool so the student sees
both formats from the start. A final stage on judged pairs, with the pairwise loss of
\Cref{eq:dpo_reranker}, sharpens its
pairwise preferences. This sequence combines the teacher's high-volume coverage
with the judges' high-precision ordering. Both released small models---the 0.6B
and the 4B---are produced this way, from their own bases. Distillation beats
aligning the same base directly at both sizes, by more on prose than on
structured text, and it also beats a costlier panel teacher built from DeepSeek
and Gemma grades: at feasible scale, label volume matters more than the nominal
strength of the teacher. \Cref{app:ablation} gives the stage-by-stage
decomposition and the paired tests.

\section{Experiments}
\label{sec:experiments}

\subsection{Experimental Setup}
\label{sec:exp_setup}

\paragraph{Benchmarks and baselines.}
The primary evaluation is the \bench preference track
(\Cref{sec:benchmark_construction}): 10,511 pairs, each in structured and
natural-language product text, with gold, silver, and bronze agreement tiers. The two
diagnostic tracks, AHP (1,500 pairs) and explicit budget (1,302 pairs), probe
constraint-following (\Cref{sec:diag_construction}). We additionally report
three public MTEB reranking tasks~\citep{muennighoff2022mteb}---%
ESCIReranking~\citep{reddy2022esci},
SciDocsRR~\citep{cohan2020specter}, and
AskUbuntuDupQuestions~\citep{lei2016askubuntu}---on which we never train.

We compare against the strongest open rerankers: Jina-m0 (2.4B), the largest
open multimodal reranker; BGE-Reranker-v2-m3 (0.6B), the size peer of our
smallest model; BGE-Reranker-large; and a BM25 lexical baseline. We also evaluate the
Qwen3-Reranker bases at all three sizes under the identical
zero-shot protocol (\Cref{lst:qwen_prompt}). These bases are not weak
references: zero-shot, base-8B already matches Jina-m0 on structured text and
edges it on prose (\Cref{tab:main_structured,tab:main_nl}). We therefore report
both comparisons throughout---against the dedicated open rerankers, and against
each model's own base, which is the comparison that isolates what alignment
adds.

\paragraph{Metrics.}
Pairwise accuracy is computed on the same pairs for every model, and we report
it separately for the three agreement tiers of
\Cref{sec:benchmark_construction}. The 10,511 pairs come from only 2,991 distinct queries, so they are not
independent observations. All intervals are therefore 95\% \emph{query-clustered}
bootstrap intervals, resampling whole query groups, and all model comparisons are
continuity-corrected paired McNemar tests~\citep{mcnemar1947note} over the same
pairs (\Cref{tab:mcnemar}). Because McNemar treats each pair as independent, we
repeated every comparison under a query-clustered bootstrap with Holm--Bonferroni
correction; no conclusion changes (\Cref{app:significance}).

\subsection{Results and Analysis}
\label{sec:results}

\begin{figure}[t]
\centering
\begin{subfigure}[t]{0.48\textwidth}
  \centering
  \includegraphics[width=\linewidth]{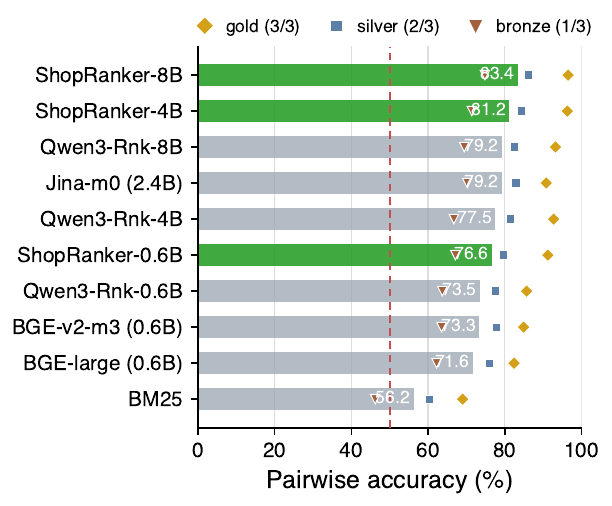}
  \caption{Structured format.}
  \label{fig:main_structured}
\end{subfigure}\hfill
\begin{subfigure}[t]{0.48\textwidth}
  \centering
  \includegraphics[width=\linewidth]{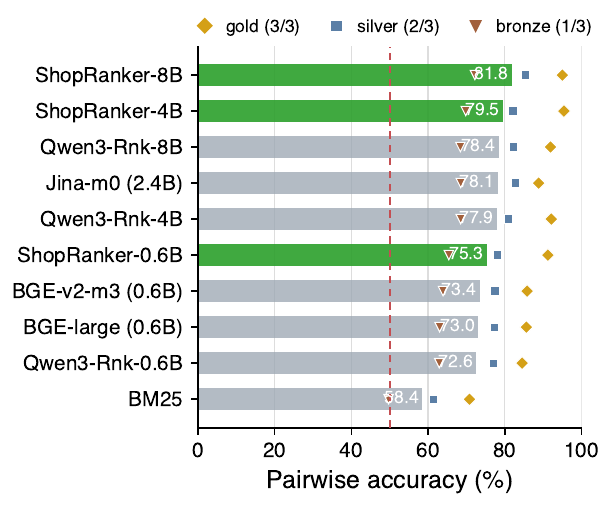}
  \caption{Natural-language format.}
  \label{fig:main_nl}
\end{subfigure}
\caption{\bench pairwise accuracy (\(n=10{,}511\)) by product-text format.
Bars show overall accuracy (\model variants in green; the un-fine-tuned
Qwen3-Reranker bases and open baselines in gray); the diamond, square, and
triangle markers show the gold, silver, and bronze agreement tiers
(\Cref{sec:benchmark_construction}). The axis starts at zero and the dashed line
marks chance (50\%), since every pair is a two-way choice. Full numbers with 95\%
query-clustered bootstrap CIs are in \Cref{app:full_results}.}
\label{fig:main_results}
\end{figure}

\paragraph{Main results.}
\model outperforms every open baseline on \bench in both product-text formats
(\Cref{fig:main_results}; full numbers in \Cref{app:full_results}). \model-8B
leads, \model-4B is a smaller alternative that still exceeds Jina-m0, and on
the natural-language gold tier even \model-0.6B beats the far larger Jina-m0
(91.2 vs.\ 88.8).
Reporting by agreement tier matters: a weaker overall score can come entirely
from the noisier single-judge bronze tier, 40\% of the benchmark, rather than
from genuine errors on well-labelled pairs. Because all
systems decide the same pairs, we judge significance with paired McNemar tests
rather than the marginal intervals, which can overlap for two systems whose
paired difference is nonetheless decisive (\Cref{tab:mcnemar}).
Rerankers are not the ceiling: two zero-shot reasoning LLMs score eight to nine
points above \model-8B on \bench{}, at seconds rather than milliseconds per
decision. A reranker is not deployed that way, so we treat this as a reference
point on how much of the benchmark's preference is recoverable at all, not as an
alternative system, and report it in \Cref{app:llm_reference}.

\begin{figure}[t]
\centering
\begin{subfigure}[t]{0.48\textwidth}
  \centering
  \includegraphics[width=\linewidth]{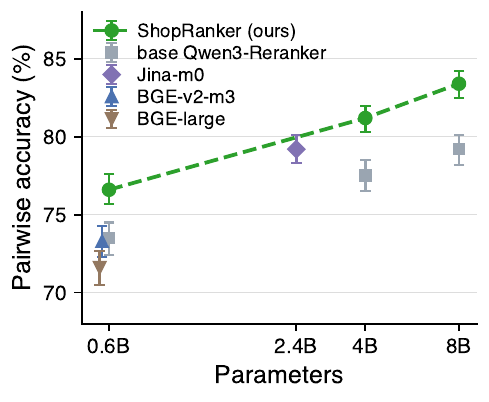}
  \caption{Structured format.}
  \label{fig:size_struct}
\end{subfigure}\hfill
\begin{subfigure}[t]{0.48\textwidth}
  \centering
  \includegraphics[width=\linewidth]{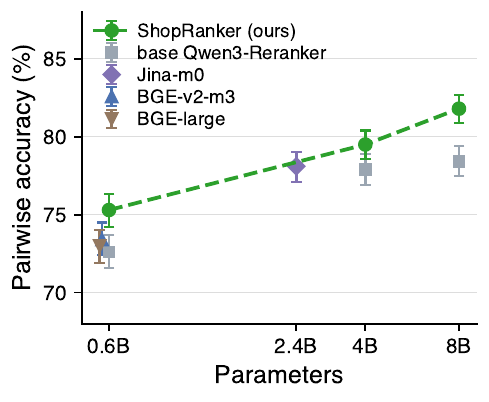}
  \caption{Natural-language format.}
  \label{fig:size_nl}
\end{subfigure}
\caption{Preference-track accuracy versus parameter count (log scale;
\(n=10{,}511\); error bars are 95\% query-clustered bootstrap CIs).
\model beats its base at every size in both formats (paired McNemar
\(p\le1.3{\times}10^{-4}\)); BM25 (56.2/58.4) is omitted for axis readability.
The size-scaling view of the two diagnostic tracks is in
\Cref{fig:size_scaling_diag} (\Cref{app:additional_analyses}).}
\label{fig:size_scaling}
\end{figure}

\paragraph{Analysis I: accuracy versus model size.}
Alignment helps at every scale, but the efficiency win is format-specific.
\Cref{fig:size_struct,fig:size_nl} plot each model against parameter count.
Every \model variant significantly improves on its Qwen3-Reranker base at all
three sizes and in both formats (paired McNemar; \Cref{tab:mcnemar}), and each
beats every open baseline of comparable or smaller size. On structured text
the distilled \model-0.6B is statistically indistinguishable from the base 4B
at a roughly seven-fold size reduction (a \(-0.9\)-point difference, 95\% CI
\([-1.9, +0.2]\); we detect no difference rather than demonstrate equality);
on natural language the
base 4B keeps a significant edge, so we report the compression result as
specific to the structured format. That reduction is what makes the pipeline
worth running: measured on one H200, the 0.6B sustains $2.8\times$ the
throughput of the 4B on $40\%$ of the memory (\Cref{app:latency}), so that
accuracy is delivered at a materially lower serving cost. Alignment helps at
every scale, then, but the deployable win is distillation.

\paragraph{Analysis II: what alignment does not cost.}
Preference alignment does not trade away general reranking quality. On three
common MTEB reranking tasks (\Cref{tab:mteb}) \model-4B and -8B lead the
average, adding a small, consistent margin over the already-strong Qwen bases,
and \model-0.6B leads its own base and its BGE size peers on that average,
though it trails BGE-v2-m3 by $0.002$ on ESCIReranking. Measured against its own
base the only slippage is SciDocsRR ($-0.008$ at 4B, $-0.003$ at 0.6B), where
fine-tuning trades a fraction for the gains elsewhere. Three tasks with no
uncertainty estimate support a check that alignment does not degrade general
reranking, not a claim of uniform improvement. We never train on any of these sets, but because public corpora may
appear in base-model pretraining, they are best read as relative comparisons
rather than uncontaminated estimates---which is precisely why \bench{}
exists. Nor do the gains come from the ablation's individual stages in
equal measure: judged-pair alignment is the dominant lever for the flagship,
distillation for the 0.6B student, and distillation adds a smaller but
significant gain over alignment from the base for the 4B, while mixed-format
training is neutral (\Cref{app:ablation}).
It costs nothing at inference either: because the LoRA update is merged before
serving, each \model{} matches its own base at the median latency
(\Cref{app:latency}). The controlled LoRA-versus-full-fine-tuning comparison and
the user-context conditioning study are in \Cref{app:additional_analyses}.
Alignment is therefore not a domain trade---the models gain e-commerce
preference without giving up general reranking quality or serving headroom.

\begin{table}[t]
\centering
\small
\setlength{\tabcolsep}{4pt}
\begin{sc}
\begin{tabular}{lcccc}
\toprule
Model & ESCIReranking & SciDocsRR & AskUbuntuDupQ & Avg. \\
\midrule
\model-4B & \textbf{0.848} & 0.882 & \textbf{0.667} & \cellcolor{green!10}\textbf{0.799} \\
\model-8B & 0.844 & 0.885 & \textbf{0.667} & \cellcolor{green!10}\textbf{0.799} \\
Qwen3-Rnk-8B & 0.835 & 0.885 & 0.659 & 0.793 \\
Qwen3-Rnk-4B & 0.836 & \textbf{0.890} & 0.641 & 0.789 \\
Jina-m0 (2.4B) & 0.837 & 0.851 & 0.632 & 0.773 \\
\model-0.6B & 0.823 & 0.835 & 0.624 & 0.761 \\
Qwen3-Rnk-0.6B & 0.816 & 0.838 & 0.610 & 0.755 \\
BGE-v2-m3 (0.6B) & 0.825 & 0.749 & 0.623 & 0.732 \\
BGE-large (0.6B) & 0.822 & 0.705 & 0.587 & 0.705 \\
\bottomrule
\end{tabular}
\end{sc}
\caption{Evaluation-only results on three common MTEB reranking tasks, each
reported at MTEB's main score for that task. No confidence intervals or
significance tests are computed for these numbers.}
\label{tab:mteb}
\end{table}

\begin{figure}[t]
\centering
\begin{subfigure}[t]{0.48\textwidth}
  \centering
  \includegraphics[width=\linewidth]{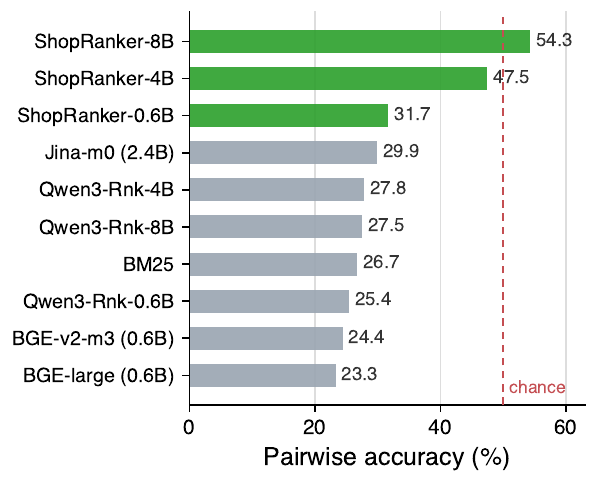}
  \caption{AHP diagnostic (\(n=1{,}500\)).}
  \label{fig:diag_ahp}
\end{subfigure}\hfill
\begin{subfigure}[t]{0.48\textwidth}
  \centering
  \includegraphics[width=\linewidth]{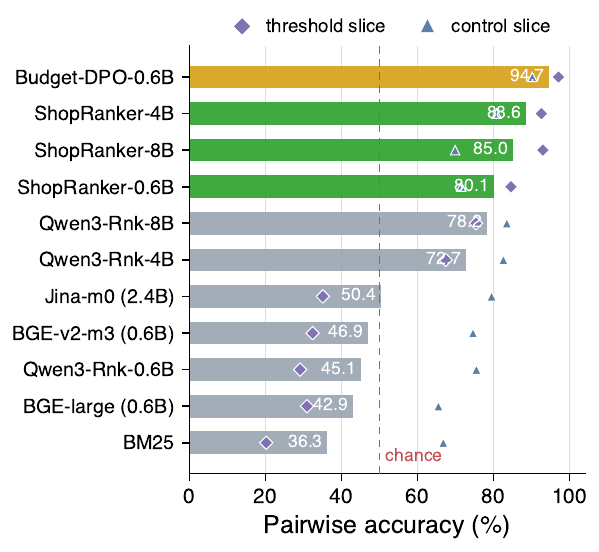}
  \caption{Explicit-budget diagnostic (\(n=1{,}302\)).}
  \label{fig:diag_budget}
\end{subfigure}
\caption{Diagnostic-track accuracy (\model in green; base Qwen3-Reranker and
open baselines in gray; the 0.6B budget specialist in gold). The dashed line is
chance (50\%): each item is a two-way choice, so scores below it mean
systematically wrong choices. On the budget track, the diamond and triangle
markers give the threshold and control slices (\Cref{sec:diag_construction}); a pick-cheaper
policy would score 100/0 on them, so only genuine threshold behavior is high on
both.}
\label{fig:diag_results}
\end{figure}

\paragraph{Analysis III: attribute hierarchy.}
\label{sec:ahp}
The AHP track asks whether a model honors a hand-designed attribute priority on
pairs that differ on exactly two attributes. Zero-shot, every open reranker and
every Qwen base sits far below the 50\% chance line (\Cref{fig:diag_ahp}): they
systematically invert the intended priority, price worst of all, and scale does
not help. Preference alignment lifts this substantially with no AHP supervision
at all, reaching just above chance at 8B. But the hierarchy is the wrong
target rather than an unmet one: a model trained to score the seven attributes
separately and pool them reaches only $38.8\%$ on judge-labeled pairs, and the
best linear re-weighting of those same attribute scores reaches $69.5\%$
held-out, so on this decomposition it is not the weights that are the binding
constraint (\Cref{app:ahp}). This bounds one learned attribute model under
linear pooling, not every possible attribute predictor or pooling function. Hand-designed attribute priority is
therefore a useful diagnostic and a poor training target.

\paragraph{Analysis IV: explicit budget.}
\label{sec:budget}
Price is the worst AHP attribute, but the failure is one of supervision rather
than capacity: soft price preference is genuinely ambiguous, whereas a
\emph{stated} budget is programmatically checkable. The budget track pairs a
threshold slice with an adversarial control slice, so neither an
always-cheaper nor an always-pricier policy can win both
(\Cref{sec:diag_construction}). The weaker open rerankers lean to the pricier
product and effectively ignore the cap (Jina-m0 and BGE-v2-m3 reach 35.1 and
32.4 on the threshold slice), while the Qwen bases do better without tracking it
(base-8B 75.4 threshold against 83.5 control); general preference alignment
overshoots toward cheap instead of learning the threshold, which is also why
budget accuracy is not monotone in size; and a 0.6B specialist trained on the
automatic rule is strong on both slices and highest overall ($94.7\%$). On this
constraint, a little targeted programmatic supervision beats both scale and
general alignment (\Cref{app:budget}); we have not tested whether that
transfers to other checkable constraints, and we do not evaluate the specialist
on the general benchmarks of \Cref{tab:mteb}. General preference
alignment shifts the price bias; it does not install budget compliance.

\section{Conclusion}
\label{sec:conclusion}

We presented \model, a family of e-commerce rerankers (0.6B--8B) aligned to
shopping preference from cross-family judge labels, with the flagship distilled
into the two smaller models. On the \bench, \model-8B and
-4B significantly outperform the strongest open baseline and \model-0.6B beats
its size peer.


\newpage
\bibliographystyle{icml2020_url}
\bibliography{reference}

\newpage
\appendix

\section{Data Preparation and Statistics}
\label{app:data}

\subsection{Category coverage of the preference set}
\label{app:categories}

Each product carries a fine-grained \texttt{product type} field, with over 900
distinct values across the 10,511 preference pairs. Bucketing these by keyword,
\bench is apparel-leaning---clothing, footwear, bags, and jewellery account for
about two-thirds of the pairs---but still spans general e-commerce: beauty,
electronics, home, baby and kids, and health together make up roughly a
quarter, and the remainder is a long tail the keyword mapping leaves
uncategorised.

We report this mix coarsely and deliberately. The benchmark is sampled from a
commercial engine's traffic, so a per-category breakdown would describe that
engine's business mix rather than any property of the benchmark. The relevant
fact for interpreting our results is the skew itself, which the proportions
above give: \bench measures e-commerce preference with an apparel emphasis, and
transfer to a catalogue weighted differently is untested.

\subsection{Judge protocol and agreement}
\label{app:judge_protocol}

Each candidate pair is judged in both presentation orders with the instruction
in \Cref{lst:judge_prompt}.

\begin{lstlisting}[style=tracebox, caption={Judge instruction. Each pair is
presented in both candidate orders and judged independently.},
label=lst:judge_prompt]
System: You are a product-preference judge. Given a query and two candidate
        products, follow these steps:
        1. State the query's hard constraints.
        2. Check each product against every hard constraint.
        3. Compare the remaining preference evidence.
        4. Decide which product is preferred, or return a tie when the
           evidence is insufficient.
\end{lstlisting}

Training retains approximately 4.4k pairs on which the Qwen3.5-122B and Gemma
families agree in both orders. Benchmark construction adds DeepSeek-V4-Pro as a
third, training-disjoint family. Of 23,000 benchmark candidates, 10,511 are
decisive (45.7\%), tiered by how many families committed: 1,843 gold (all
three), 4,445 silver (two, one abstaining), and 4,223 bronze (one, two
abstaining), as set out in \Cref{sec:benchmark_construction}. Unanimous ties and
outright conflicts are discarded.

\subsection{Held-out judge audit}
\label{app:fourth_judge}

The preference labels come from the three-family panel of
\Cref{app:judge_protocol} and are not human-verified. As a check that the panel
is not encoding a single-family artefact, we audited a tier-stratified sample of
45 released pairs (15 per tier) with Claude Opus 5, a fourth model family
held out of both training and benchmark construction and distinct from the
reference LLMs of \Cref{app:llm_reference}. The adjudicator saw only the query and the
two product texts, with the candidate order randomised independently per pair,
and committed every verdict before any label was revealed. Agreement with the
released label was 43/45: $15/15$ gold, $15/15$ silver, and $13/15$ bronze. Both
disagreements fell in bronze---the tier where a single family committed a
verdict---and both had been marked low-confidence by the adjudicator at decision
time, one of them a query (``christmas scarf'') on which a red silk scarf and a
tartan one are not obviously separable. With 15 pairs per tier the intervals are
wide (total $95.6\%$, $95\%$ Wilson $[85.2, 98.8]$), so this is a consistency
check rather than an estimate of label accuracy, and it is not human validation.

\subsection{Does the margin depend on the training judges?}
\label{app:circularity}

Two of the three benchmark families, Qwen3.5-122B and Gemma-4-31B, also produced
the training labels; only DeepSeek-V4-Pro is training-disjoint
(\Cref{sec:train_setup}). If our gains merely reproduced the training judges'
idiosyncrasies, they should shrink on pairs those judges did not decide. We
therefore split the non-gold pairs by whether DeepSeek committed a verdict:
5,100 pairs where it did, and 3,568 where it abstained and the label came from
the training families alone.

\begin{table}[h]
\centering
\small
\setlength{\tabcolsep}{5pt}
\begin{sc}
\begin{tabular}{lccc}
\toprule
& & \multicolumn{2}{c}{Non-gold pairs} \\
\cmidrule(lr){3-4}
Model & \makecell{Gold\\($n{=}1{,}843$)} & \makecell{DeepSeek voted\\($n{=}5{,}100$)}
 & \makecell{Training panel only\\($n{=}3{,}568$)} \\
\midrule
\model-8B & 96.5 & 83.6 & 76.2 \\
\model-4B & 96.3 & 81.6 & 72.8 \\
Qwen3-Rnk-8B & 93.2 & 81.0 & 69.3 \\
Jina-m0 (2.4B) & 90.8 & 80.7 & 71.0 \\
\model-0.6B & 91.2 & 77.4 & 68.1 \\
\midrule
\multicolumn{4}{l}{\emph{Advantage over Jina-m0}} \\
\model-8B & $+5.7$ & $+2.9$ & $+5.2$ \\
\model-4B & $+5.5$ & $+0.9$ & $+1.8$ \\
\bottomrule
\end{tabular}
\end{sc}
\caption{Accuracy on \bench{} structured pairs, split by whether the
training-disjoint judge family committed a verdict. The \model{} advantage
survives on pairs that DeepSeek also decided ($+2.9$ and $+0.9$), which is the
subset that bears on circularity. The ``DeepSeek voted'' column is not
training-disjoint---the training families may also have voted on those
pairs---so this bounds a training-family effect rather than removing it.}
\label{tab:circularity}
\end{table}

\Cref{tab:circularity} shows the advantage over Jina-m0 persisting on both
subsets. The subset that bears on circularity is the one a training-disjoint
family helped decide, and there the advantage survives but shrinks---from $+5.2$
to $+2.9$ for the 8B and from $+1.8$ to $+0.9$ for the 4B. We read this as
bounding a training-family effect rather than excluding one: the advantage is
\emph{larger} where only the training families committed, which is what partial
fit to those families would also produce. Three further caveats apply. The
``DeepSeek voted'' column does not exclude pairs the training families also
decided, so the two columns are not cleanly disjoint; their tier composition
differs; and we report no interaction test. The absolute accuracies fall on the
training-panel-only subset for every system alike, which is the label-strength
effect of \Cref{sec:benchmark_construction} rather than a circularity effect.
What we can say is that the gain is not \emph{only} an artefact of matching the
training judges, not that it is free of one. A human-agreement study would test
the stronger claim, that the panel tracks shopper preference at all, and we
leave it as the main open item.

\subsection{Contamination audit}
\label{app:contamination}

Two properties support the contamination-limited claim of \Cref{sec:data}, in the
sense defined there. First, the candidate pairings and their labels are built
from private traffic that has never been published, so no base model can have
seen the answers; this is a property of the source rather than a measurement,
and it does not extend to the individual queries or to the public catalogue
attributes the product text renders from, neither of which we can audit against
any pretraining corpus. Second, the benchmark is disjoint from our own
training corpus: the training slice and \bench{} are drawn from
non-overlapping traffic windows, and a direct check finds zero shared query
strings and zero shared (query, positive, negative) triples between them.
\bench{} is also internally non-redundant: across the 10,511 released
records no (query, preferred, rejected) triple occurs twice, no candidate
pair recurs with its label reversed, and no record pairs a product against
itself. Controlling the source of evaluation data in this way is likewise
central to recent domain benchmarks in financial multimodal question
answering~\citep{xue2024famma} and time-series
forecasting~\citep{xue2026quitobench}.

We additionally verify that the benchmark is not solvable by surface artifacts
of its construction, reporting each heuristic in its stronger direction.
Always picking the longer product text wins $5{,}602$ of the $10{,}511$
released preference pairs ($53.3\%$); $153$ pairs carry two texts of equal
length, so no tie-breaking rule lifts this above $54.1\%$. Both products carry
a parsable price in every record, and $339$ pairs price them equally; on the
remaining $10{,}172$ pairs, always picking the cheaper product wins $5{,}969$
($58.7\%$, the pricier $4{,}203$ or $41.3\%$)---above BM25's $56.2\%$, though
BM25 is scored on the full pair set rather than on this subset---but still far
below every learned model in \Cref{tab:main_structured}, the weakest of which
scores $71.6\%$. Both audits are reproduced by the released evaluation
code. A benchmark
that a length or price heuristic could win would measure formatting
regularities rather than preference. That pick-cheaper is above chance is the
organic prefer-cheaper signal of \Cref{sec:intro}: among two comparably
relevant products the judges more often favour the cheaper one, which is a
property of shopper-facing preference rather than an artefact of construction.
A related property is visible in the candidate pool rather than the labels.
Because candidates are retrieved rather than curated, a minority of pairs are
separated by product type instead of preference---a food container retrieved
against a refrigerator query, an accessory against the device it fits---and some
catalogue attributes are themselves noisy, so a snowboard boot may be typed as a
shoe. These pairs are correctly labelled but easy, and they are part of why the
aligned models reach the mid-$90$s on the gold tier in
\Cref{tab:main_structured} while separating far more on silver and bronze.
Gold is not uniformly easy---it spans $69.0\%$ (BM25) to $96.5\%$ across the
systems in that table---but it is where the strong rerankers bunch together.

\subsection{Data release and privacy}
\label{app:privacy}

\bench{} is derived from logged search traffic, so we describe what the
released records contain and why they carry no personal information. Each preference-track
record carries only the query string, the two product texts, the agreement
tier, and the per-judge verdicts (\Cref{lst:pref_example}); diagnostic records
add only the construction fields shown in
\Cref{lst:ahp_example,lst:budget_example}. No user, account, session, device,
location, timestamp, or behavioural field (clicks, purchases, dwell) is
released, and queries are not linked to one another by user or session, so a
user's search history cannot be reconstructed from the release. Product text is
rendered from public catalogue attributes (\Cref{app:schema}) and contains no
user content.
The queries themselves are product searches---what a shopper is looking for,
such as a product type, attribute, occasion, or budget---and contain no
personally identifying information. We screened all $2{,}991$ distinct released
queries for direct identifiers---email addresses, telephone numbers, street
addresses, postal codes, government identifiers, long digit strings, social
handles, and URLs---and found none. Where a query mentions a relationship or a
use context (e.g., ``printer for my moms house''), it describes the product need
and does not identify a person; $35$ queries do so. \bench{} is distributed for research use.

\subsection{Natural-language rendering styles}
\label{app:nl_styles}

The dual-format release uses five rendering styles that vary sentence order,
attribute selection, and phrasing, so no single prose template can be
memorized. Each rendering is produced by a local LLM from the structured fields
only; no scraped description is introduced, and the same preference label
applies to every rendering.

\subsection{Product text schema}
\label{app:schema}

The structured view uses the canonical pipe-delimited product representation.
Its 16 fields are retained below.

\begin{table}[h]
\centering
\small
\begin{sc}
\begin{tabular}{ll}
\toprule
Field & Example \\
\midrule
style & casual, sporty \\
color & black \\
material & cotton \\
audience & women \\
product type & t-shirt \\
brand & Zara \\
pattern & solid \\
fit & regular \\
length & midi \\
neckline & crew \\
closure & zipper \\
sleeve & long \\
function & waterproof \\
occasion & casual \\
season & spring \\
price & 29.99 \\
\bottomrule
\end{tabular}
\end{sc}
\caption{Product text schema fields.}
\label{tab:schema}
\end{table}

\begin{lstlisting}[style=productjson,caption={Example structured product text.}]
{Brand} {ProductType}
style: casual | color: black | material: cotton |
audience: women | product type: t-shirt |
brand: Zara | pattern: solid | fit: regular |
price: 29.99 | ...
\end{lstlisting}

\subsection{Track record examples}
\label{app:track_examples}

\Cref{lst:pref_example,lst:ahp_example,lst:budget_example} show one record
from each \bench track; the dual-format view of a preference-track product is
in \Cref{lst:dual_format} in the main text. Product texts follow the structured schema
of \Cref{app:schema}; long attribute lists are elided with ``...'' for
presentation, and the released files contain the full records.

\begin{lstlisting}[style=productjson,caption={Preference-track record. The
query is a real user search string; \texttt{pos} is the judge-preferred
product. Gold tier means all three judge families agree in both presentation
orders.},label=lst:pref_example]
{
 "query": "sonic socks for boys 4-6 years",
 "pos_product_text": "style: ... | color: ... |
    audience: ... | product type: socks | ... |
    price: ... | product title: ...",
 "neg_product_text": "style: ... | ... ",
 "tier": "gold",
 "judges": ["Qwen3.5-122B", "Gemma-4-31B",
            "DeepSeek-V4-Pro"],
 "consistent_both_orders": true
}
\end{lstlisting}

\begin{figure}[h]
\begin{minipage}[t]{0.48\textwidth}
\begin{lstlisting}[style=productjson,basicstyle={\codefont\footnotesize},caption={AHP diagnostic record, verbatim from the released track. The two
products differ on exactly two hierarchy attributes and are identical on every
other catalogue field: \texttt{pos} satisfies the higher-priority one (price,
given the stated \$30 cap) while violating the lower-priority one (color), and
\texttt{neg} the reverse; the hierarchy label picks \texttt{pos}.},label=lst:ahp_example]
{
 "query": "looking for watch
    under $30 brown",
 "query_constraints": {
    "price": "under $30",
    "color": "brown"},
 "pos_product_text": "style: casual
    | color: purple | material:
    silicone | audience: kids |
    product type: watch | brand:
    OLEVS | ... | price: 25 |
    product title: OLEVS Watch",
 "neg_product_text": "style: casual
    | color: brown | material:
    silicone | audience: kids |
    product type: watch | brand:
    OLEVS | ... | price: 45 |
    product title: OLEVS Watch",
 "ahp_level_high": "price",
 "ahp_level_low": "color",
 "pos_satisfies": "price",
 "neg_satisfies": "color",
 "expected_winner": "pos"
}
\end{lstlisting}
\end{minipage}\hfill
\begin{minipage}[t]{0.48\textwidth}
\begin{lstlisting}[style=productjson,basicstyle={\codefont\footnotesize},caption={Explicit-budget record
(\emph{threshold} slice): one product meets the stated \$40 budget and the
other exceeds it, so the programmatic oracle
$\mathrm{price}\le\mathrm{budget}$ decides. In the \emph{control} slice the
budget is raised above both prices and the pricier product wins by
construction.},label=lst:budget_example]
{
 "query": "affordable black womens
    pullover hoodie under $40",
 "pos_text": "style: casual | color:
    caviar | audience: women |
    product type: hoodie | brand:
    Reef | ... | price: 34.99 |
    product title: Reef hoodie",
 "neg_text": "style: casual,
    athleisure | color: harley black
    | audience: women | product type:
    hoodie | brand: Harley-Davidson |
    ... | price: 90.0 |
    product title:
    Harley-Davidson hoodie",
 "slice": "price_threshold",
 "budget": 40.0,
 "winner_price": 34.99,
 "loser_price": 90.0
}
\end{lstlisting}
\end{minipage}
\end{figure}

\section{Additional Analyses}
\label{app:additional_analyses}

\subsection{Training hyperparameters}
\label{app:hparams}

Every stage trains a LoRA adapter of rank $16$ ($\alpha=32$, dropout $0.05$) on
the attention projections (\texttt{q,k,v,o\_proj}) of each layer, with the base
weights frozen (\Cref{sec:train_setup}). Shared across stages: maximum sequence
length $1024$, effective batch $64$, gradient-norm clipping at $1.0$, weight
decay $0.01$, linear warmup ($0.10$; $0.05$ for the distillation passes) then
linear decay, fp16 with gradient checkpointing, and best-checkpoint selection on
the held-out judged validation split by pairwise accuracy. The learning rate is
$5\times10^{-6}$ for the judged-pairs, distillation and budget-specialist
stages and $2\times10^{-6}$ for the mixed-format and pairwise-sharpening
stages; the pairwise logistic loss of \Cref{eq:dpo_reranker} uses $\beta=0.1$,
and distillation fits the 8B teacher's scores with BCE. The superseded
alignment-from-base runs of \Cref{tab:ablation} share these settings.

Each stage samples over its data sources rather than capping any realised batch:
judge-labeled e-commerce pairs are drawn with probability $0.60$ in the
judged-pairs stage and $0.80$ in the mixed-format and sharpening stages, the
remainder being the general-relevance replay anchor of \Cref{sec:dpo}.

\subsection{Diagnostic tracks by model size}
\label{app:size_scaling_diag}

\begin{figure}[h]
\centering
\begin{subfigure}[t]{0.48\textwidth}
  \centering
  \includegraphics[width=\linewidth]{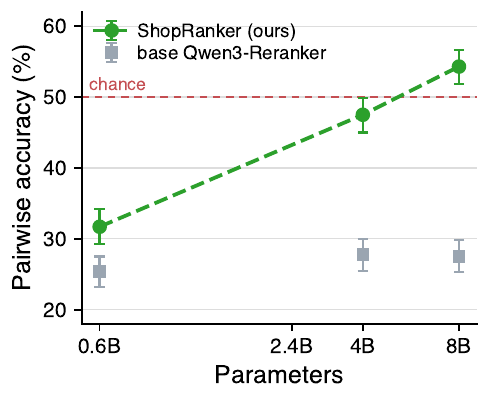}
  \caption{AHP diagnostic (\(n=1{,}500\)).}
  \label{fig:size_ahp}
\end{subfigure}\hfill
\begin{subfigure}[t]{0.48\textwidth}
  \centering
  \includegraphics[width=\linewidth]{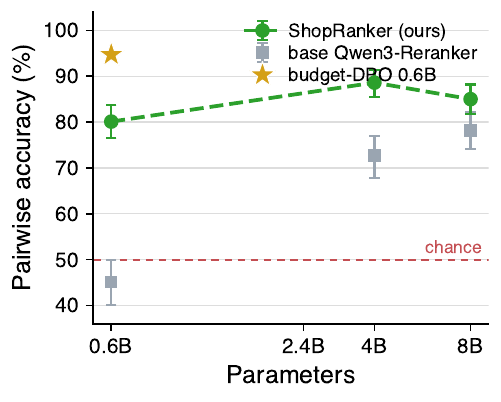}
  \caption{Explicit-budget diagnostic (\(n=1{,}302\)).}
  \label{fig:size_budget}
\end{subfigure}
\caption{Diagnostic-track accuracy versus parameter count (log scale). The
dashed line is chance (50\%).}
\label{fig:size_scaling_diag}
\end{figure}

\Cref{fig:size_scaling_diag} is the size-scaling counterpart of the ranked bars
in \Cref{fig:diag_results}, and it separates two behaviours that the bars
conflate. On AHP, the un-fine-tuned Qwen3-Reranker bases are flat and far below
chance at every size (25.4, 27.8, 27.5 at 0.6B, 4B, 8B), so scale alone does not
recover the hierarchy. Preference alignment, which contains no AHP supervision,
lifts adherence monotonically with size (31.7, 47.5, 54.3), reaching just above
chance only at 8B; \Cref{app:ahp_breakdown} shows this lift is a redistribution
across attributes rather than a uniform gain. On the budget track the pattern is
different: alignment improves compliance over the base at every size, but the
curve is not monotone---\model-8B trails \model-4B because it overshoots toward
the cheaper product on the control slice (\Cref{app:budget})---and the 0.6B
budget specialist stays ahead of every general model regardless of size.

\subsection{Constraint-subset breakdown}
\label{app:constraint_breakdown}

\Cref{tab:constraint_breakdown} gives the per-subset numbers behind
\Cref{fig:constraint_gap}, including the budget slice that the figure omits for
legibility. Constraint types are detected from the query by pattern matching; the three
constraint rows overlap slightly, since one query can state both a budget and a
recipient.

The budget row carries the sharpest result in the table. On the 338 gold pairs
that name a price ceiling, BM25 falls below chance (47.9) and the cross-encoders,
while staying above it, drop to 55.6--70.4 from 89.6--96.1 on pairs that state no
constraint, because a relevance ranker has no reason to prefer the cheaper of two
equally relevant products, whereas \model-4B reaches 97.9. The exclusion subset
is small ($n=48$) and its numbers should be read as indicative only.

\begin{table}[h]
\centering
\small
\setlength{\tabcolsep}{3pt}
\begin{sc}
\begin{tabular}{lrcccccccc}
\toprule
Subset & $n$ & BM25 & \makecell{BGE\\lg} & \makecell{BGE\\v2-m3} & \makecell{Jina\\m0}
 & \makecell{Base\\8B} & \makecell{Ours\\0.6B} & \makecell{Ours\\4B} & \makecell{Ours\\8B} \\
\midrule
All gold & 1,843 & 69.0 & 82.4 & 84.9 & 90.8 & 93.2 & 91.2 & 96.3 & \textbf{96.5} \\
\addlinespace[2pt]
No stated constraint & 1,159 & 75.1 & 89.6 & 90.3 & \textbf{96.1} & 95.3 & 91.1 & 95.9 & 96.3 \\
Any hard constraint & 684 & 58.6 & 70.0 & 75.6 & 81.9 & 89.8 & 91.4 & \textbf{96.9} & 96.8 \\
\addlinespace[2pt]
\quad budget & 338 & \cellcolor{red!10}47.9 & 55.6 & 63.9 & 70.4 & 89.9 & 92.3 & \textbf{97.9} & \textbf{97.9} \\
\quad recipient & 329 & 71.4 & 82.4 & 86.3 & 92.4 & 91.5 & 91.2 & \textbf{96.0} & \textbf{96.0} \\
\quad exclusion & 48 & \cellcolor{red!10}45.8 & 64.6 & 70.8 & 72.9 & 68.8 & 83.3 & \textbf{97.9} & 91.7 \\
\bottomrule
\end{tabular}
\end{sc}
\caption{Gold-tier accuracy by constraint subset. Shaded cells are at or below
the 50\% chance line. ``Base-8B'' is the un-fine-tuned Qwen3-Reranker-8B.
The three indented rows are subsets of ``any hard constraint'' and overlap.}
\label{tab:constraint_breakdown}
\end{table}

\subsection{Pipeline ablation}
\label{app:ablation}

Within the alignment-from-base route, every stage except mixed-format training
contributes, and which stage matters most depends on size:
distillation is the dominant lever for the 0.6B student, judged-pair alignment for the 8B,
and on-policy refinement adds a smaller increment for the 4B. Mixed-format
training is close to neutral on these development slices---hard-NL moves by
$+0.6$ at 4B, $+0.3$ at 8B, and $-0.2$ at 0.6B, and the 8B's hard versus
hard-NL gap is 3.8 points before the stage and 4.0 after.

Two of these checkpoints were also evaluated on \bench{} itself, to settle the
questions the development slices leave open. The final alignment-from-base 4B
is significantly weaker than the released distilled \model-4B in both formats
($80.4$ vs.\ $81.2$ structured, McNemar \(\chi^2=7.6,\ p=0.006\); $78.3$ vs.\
$79.5$ prose, \(\chi^2=12.7,\ p=3.7{\times}10^{-4}\)), so distillation, not only
the 0.6B evidence above, justifies the 4B's route. The structured-only 8B is indistinguishable from the
released mixed-format \model-8B in both formats, so the mixed-format stage is
neutral on the benchmark as well (\Cref{app:format_training}).

\paragraph{Panel teacher.} We also tested a teacher built from DeepSeek and
Gemma grades rather than the aligned 8B. Its labels were $95.5\%$
pairwise-consistent in validation with only $0.5\%$ reversals, but it
underperformed at feasible scale: on the development hard-structured slice
(\(n=1{,}910\)) the panel-distilled \model-0.6B reaches 70.7 against 78.3 for
the same student distilled from the reranker teacher. The panel grades far fewer
examples per unit of compute, and the coverage lost outweighs the precision
gained.

On these development slices the full pipeline lifts every model above the
next-larger base model on hard pairs, and the distilled 0.6B reaches the
base-8B level ($78.3$ vs.\ $78.0$ hard). That is a development-set statement;
the corresponding claim on \bench{} itself is the narrower one of Analysis~I,
where the 0.6B is statistically indistinguishable from base-4B on structured
text only (\Cref{tab:mcnemar}).
Together with the negative results reported alongside each stage, this shows
the gains come from the recipe, not from scale or full parameter updates alone.
\Cref{tab:ablation} gives the stage-by-stage decomposition.

\begin{table}[h]
\centering
\small
\setlength{\tabcolsep}{3pt}
\begin{sc}
\begin{tabular}{llccc}
\toprule
Size & Stage & Easy & Hard & Hard-NL \\
\midrule
0.6B & Base & 77.7 & 72.6 & 71.2 \\
 & + judged pairs & 78.3 & 73.9 & 71.6 \\
 & + distill (4B teacher) & 82.5 & 78.3 & 73.8 \\
 & + mixed format & 82.5 & 78.3 & 73.6 \\
\midrule
4B & Base & 82.1 & 76.4 & 75.8 \\
 & + judged pairs & 83.6 & 78.7 & 75.7 \\
 & + on-policy & 84.4 & 80.0 & 75.7 \\
 & + mixed format & 84.5 & 79.8 & 76.3 \\
\midrule
8B & Base & 83.5 & 78.0 & 75.1 \\
 & + judged pairs & 87.2 & 82.3 & 78.5 \\
 & + mixed format & 87.3 & 82.8 & 78.8 \\
\bottomrule
\end{tabular}
\end{sc}
\caption{Ablation of the alignment-from-base pipeline on the two-judge
development slices of \Cref{sec:train_setup} (easy, hard, hard-NL). This is the
route the released 0.6B and 4B do \emph{not} take---they are distilled from the
aligned 8B instead (\Cref{sec:distillation})---so the table motivates the recipe
rather than decomposing the shipped models. Rows are consistent within this
table and are not directly comparable to the three-judge release benchmark.
The 0.6B distillation rows here use an earlier 4B teacher, superseded by the 8B
teacher of the released models.}
\label{tab:ablation}
\end{table}

\subsection{Mixed-format training}
\label{app:format_training}

On 500 hard development pairs rendered in both formats, the structured-only
aligned 8B lost 4.6 points when the same products were written as prose (84.2
to 79.6), against 3.0 for base-4B (76.6 to 73.6), which suggested that alignment
on the structured schema makes a model brittle on prose. We therefore added a
final pass mixing the structured schema with model-rendered prose. A controlled
comparison on \bench{} refutes the fix and leaves the premise untested:
the structured-only 8B and the released
mixed-format \model-8B are indistinguishable in both formats (83.3 vs.\ 83.4
structured, McNemar \(p=0.94\); 81.7 vs.\ 81.8 prose, \(p=0.64\)), and both
have the same 1.6-point format gap. Alignment on structured text alone already
carries its gain to prose, and the development-set brittleness does not
reproduce at benchmark scale. Whether alignment \emph{increases} format
sensitivity is a separate question these tests do not answer: they compare
checkpoints within each format rather than the difference between format gaps,
and the aligned 8B's 1.6-point gap against base-8B's 0.8
(\Cref{tab:main_structured,tab:main_nl}) would need an interaction test we do
not run. We keep the stage in the released recipe, since it
costs nothing, but claim no robustness benefit for it.

\subsection{Serving cost}
\label{app:latency}

Latency is largely a property of the base model rather than of our recipe: the
LoRA update is merged into the weights before serving, so each \model{} costs
what its Qwen3-Reranker base costs. \Cref{tab:latency} records that, and two
consequences worth stating.

First, alignment is free at inference. Each \model{} matches its own base at the
median at every size (17.6 vs.\ 18.6\,ms at 0.6B, 25.2 vs.\ 24.5 at 4B, 25.8
vs.\ 25.4 at 8B), so the accuracy gains of \Cref{fig:main_results} cost nothing
per query. Second, and this one is ours rather than the base model's, the
distilled \model-0.6B is statistically indistinguishable from base-4B on
structured \bench{} pairs (Analysis~I) while sustaining $2.8\times$ the
throughput (433 vs.\ 155 docs/s) on $40\%$ of the memory (4.5 vs.\ 11.2\,GB).
The 8B flagship buys a further 6.8 accuracy points over the 0.6B at $4.2\times$
the serving cost---a trade a deployment can now price. Batching dominates
throughput: a single 0.6B request takes 17.6\,ms at the median, whereas at batch
32 its 433 docs/s amortise to a mean of 2.3\,ms per document, so the batch-1
columns are a latency budget, not a throughput figure.

We also report the comparison that does not favour us. BGE-v2-m3, the encoder
cross-encoder we beat by 3.3 points on structured \bench{}, is $4.6\times$
faster at batch 32 and needs a third of the memory. A decoder reranker scoring a
yes/no token carries the full language-model stack to produce one scalar; an
encoder does not. Where quality binds, our models are the better choice; where
throughput binds, the encoder still wins. At the other end, the zero-shot LLMs
of \Cref{app:llm_reference} are 8.5 to 8.7 points more accurate than \model-8B but
take a median of 3.4 to 11.8\,s per call and \$3.6 per thousand pairs, so
no decoder reranker closes that gap at serving-time latency.

\begin{table}[h]
\centering
\small
\setlength{\tabcolsep}{4pt}
\begin{sc}
\begin{tabular}{lccccc}
\toprule
Model & Params & \makecell{Batch-1\\p50 (ms)} & \makecell{Batch-1\\p95 (ms)} & \makecell{Docs/s\\(batch 32)} & \makecell{Peak\\mem (GB)} \\
\midrule
\model-0.6B & 0.60B & 17.6 & 25.1 & 433.1 & 4.5 \\
\model-4B & 4.02B & 25.2 & 33.3 & 154.5 & 11.2 \\
\model-8B & 8.19B & 25.8 & 30.4 & 102.7 & 18.9 \\
\midrule
Qwen3-Rnk-0.6B & 0.60B & 18.6 & 20.3 & 431.9 & 4.5 \\
Qwen3-Rnk-4B & 4.02B & 24.5 & 26.6 & 154.9 & 11.2 \\
Qwen3-Rnk-8B & 8.19B & 25.4 & 28.4 & 102.8 & 18.9 \\
Jina-m0 (2.4B) & 2.21B & 17.3 & 26.2 & ---$^\dagger$ & 4.2 \\
BGE-v2-m3 (0.6B) & 0.57B & \textbf{6.7} & \textbf{10.4} & \textbf{1998.5} & \textbf{1.3} \\
\bottomrule
\end{tabular}
\end{sc}
\caption{Serving cost on one NVIDIA H200, fp16, 1024-token limit, measured over
256 real \bench{} (query, document) pairs after 8 untimed warmup batches, with
the device synchronised around every timed region. Batch-1 latency is the
serving-budget number and is measured identically for every system;
throughput is reported only where a batched forward pass exists.
$^\dagger$Jina-m0 is driven through its upstream per-pair API, so it has no
comparable batched figure and is not given one. Parameter counts are of the
merged serving model as loaded; Jina-m0's nominal size is 2.4B, the figure we
use to label it elsewhere, and 2.21B is the count of its loaded weights. The GPU was exclusive to this job but the host was shared
with an unrelated CPU-only workload, which widens the p95 tail; the p50 column
is the stable comparison. BGE-Reranker-large was not evaluated here.}
\label{tab:latency}
\end{table}

\subsection{Zero-shot LLM reference}
\label{app:llm_reference}

Because the labels come from LLM judges, a natural question is how well an LLM
does as the ranker itself. We evaluate two reasoning LLMs, GLM-5.3 and
Qwen3.5-27B, on all 10,511 structured \bench{} pairs. GLM-5.3 is from a family
outside the judge panel. Qwen3.5-27B is not: it is a much smaller model from one
of the judging families, so its agreement with the labels is not independent of
the supervision and we report it as a scale reference rather than as evidence
that the labels generalise across families. The prompt is a plain forced choice (``which product better matches what
the shopper wants? Answer A or B''), deliberately not the constraint-first judge
protocol of \Cref{lst:judge_prompt}, and each pair is asked in both presentation
orders at temperature 0 with the models' default reasoning. A pair counts as
correct only when both orders pick the preferred product; counting an
order-split pair as half correct, which is the expected accuracy of a single
randomly ordered call, gives 94.7\% and 94.9\%. Both models return a parseable
answer on at least 99.8\% of calls and agree with themselves across orders on 94\% of
pairs.

\begin{table}[h]
\centering
\small
\setlength{\tabcolsep}{4pt}
\begin{sc}
\begin{tabular}{lcccccc}
\toprule
Model & Overall & Gold & Silver & Bronze & \makecell{p50 per\\call} & \makecell{\$ per\\1k pairs} \\
\midrule
GLM-5.3 & 91.9 & 99.6 & 95.9 & 84.3 & 3.4\,s & 3.56 \\
Qwen3.5-27B & 92.1 & 99.9 & 96.0 & 84.6 & 11.8\,s & 3.67 \\
\model-8B & 83.4 & 96.5 & 86.1 & 74.8 & 25.8\,ms & --- \\
\bottomrule
\end{tabular}
\end{sc}
\caption{Zero-shot LLM reference on structured \bench{} (\(n=10{,}511\)). LLM
latency is API wall-clock per call---GLM-5.3 through Fireworks, Qwen3.5-27B
through OpenRouter---with mean output of 315 (GLM) and 1,140 (Qwen) tokens,
almost all reasoning; cost covers both orders at list price. Both timing and
cost are route-dependent and are not properties of the models themselves. The \model-8B latency is batch-1 p50 on one H200
(\Cref{tab:latency}). McNemar against \model-8B: \(\chi^2=416,\
p=1.6{\times}10^{-92}\) (GLM) and \(\chi^2=453,\ p=1.6{\times}10^{-100}\)
(Qwen); the two LLMs do not differ (\(p=0.46\)).}
\label{tab:llm_reference}
\end{table}

Both LLMs clearly beat every reranker (\Cref{tab:llm_reference}), and they do
not differ from each other (\(p=0.46\)), so the out-of-panel GLM-5.3 carries the
reading below and Qwen3.5-27B corroborates it. On gold
pairs both sides are near the ceiling and the lead is about three points; on
silver and bronze it is about ten. Two readings follow. First, the preference the
benchmark encodes is largely recoverable by a capable reasoning LLM outside the
judge panel, without the judge protocol or any fine-tuning, so the gap to
\model{} is a limit of the reranker, not noise in the labels. Second, these LLMs are not the judges but
belong to the same class of models, so part of their lead may be shared LLM
judgement rather than shopper preference; this is the circularity caveat of
\Cref{app:circularity} in a stronger form, and only behavioural signals can
separate the two. The cost gap is real but smaller than a
generated rationale suggests: scored by parallel decoding rather than made to
write out its reasoning, a 27B decides a pair in 224\,ms, against the
$2\times25.8$\,ms of the two \model-8B scores that the same pairwise decision
requires---a factor of about four before batching, and a reranker must score tens
of candidates per query where the LLM answers once. \model{} therefore recovers
most of this preference at a fraction of the serving cost, rather than the LLM
being the only affordable option. That 224\,ms comes from a
separate 2,000-pair run, not from the full-benchmark evaluation in
\Cref{tab:llm_reference}, whose accuracies are measured with generated reasoning
at 3.4 and 11.8\,s per call; we therefore quote it as a cost figure and make no
accuracy claim for that decoding mode at benchmark scale. Distilling LLM-level preference
into a reranker is thus an open target rather than a solved one, and
\bench{} now measures the distance to it.

\subsection{Parameter efficiency}
\label{sec:param_efficiency}

\Cref{tab:efficiency} isolates the contribution of the adapter-only recipe.

\begin{table}[h]
\centering
\small
\setlength{\tabcolsep}{4pt}
\begin{sc}
\begin{tabular}{lccc}
\toprule
Method & Easy & Hard & Hard-NL \\
\midrule
LoRA (0.6B, controlled) & \textbf{78.3} & \textbf{73.9} & \textbf{71.6} \\
Full FT (0.6B, controlled) & 77.7 & 73.0 & 71.0 \\
\bottomrule
\end{tabular}
\end{sc}
\caption{Controlled LoRA versus full fine-tuning at 0.6B: identical data,
recipe, and evaluation, differing only in whether the update is low-rank.
Measured on the development slices of \Cref{sec:train_setup}.}
\label{tab:efficiency}
\end{table}

Under the controlled 0.6B recipe, LoRA matches or edges full fine-tuning while
updating approximately 1\% of parameters. The resulting adapter is a few
megabytes rather than a multi-gigabyte checkpoint. Because the two arms differ
only in the update's rank, the comparison isolates that choice: the gains
reported elsewhere in this paper come from the data and the training objective,
not from updating every parameter.

\subsection{User-context conditioning}
\label{sec:robustness}

We test 240 synthetic cases: explicit-gender control, profile-versus-explicit
conflict, profile disambiguation, and purchase-history transfer, with 60 cases
of each type. All models are 100\% accurate, so these cases separate no system
on accuracy and we report decision margins instead. On the hard case types,
base-4B margins are 0.25/0.30/0.68, whereas \model-8B is approximately 1.0
throughout. That comparison changes both scale and training, so it does not
isolate an effect of fine-tuning, and we have no calibration evidence that a
larger margin implies a more robust decision; we report it as a description of
these 240 cases rather than as a gain.
Crucially, explicit query intent overrides a conflicting user profile.

\subsection{Attribute hierarchy: full analysis}
\label{app:ahp}

The AHP track is a controlled test of \emph{constraint priority}. Each pair
differs on exactly two attributes of a hand-designed importance ranking---for
example, matching the intended audience is treated as more important than
matching a requested color---and the model should prefer the product that
satisfies the higher-priority attribute. Because every item is a two-way
choice, 50\% is the random-guess baseline (the ``chance'' line in
\Cref{fig:diag_ahp}); scoring below it means a model systematically inverts the
intended priority. Zero-shot, every open reranker and every un-fine-tuned Qwen
base sits well below chance (\Cref{fig:diag_ahp}; per-attribute numbers in
\Cref{tab:ahp_results}), price is the hardest attribute, and scale does not
help.

We do \emph{not} use AHP as a training target, because the hand-designed
priorities disagree with judged preference, and the disagreement is large. A
multi-head model trained to score the seven attributes separately and combine
them reaches only $38.8\%$ on the judge-labeled development slice
(\(n=5{,}076\))---below the $50\%$ chance line, meaning the hierarchy actively
inverts judged preference---against $82.5\%$ for the relevance-trained 0.6B on
that same slice. Nor is this merely a badly-chosen weighting: fitting the
best linear pooling of those attribute scores directly against the
judge, which is the best any \emph{linear} re-weighting of these particular
scores can do, reaches $69.5\%$ held-out. It is these attribute scores, not the weights
placed on them, that fail to capture judged preference. Notably, these
priorities are already intent-conditioned rather than
a single global ranking (\Cref{sec:diag_construction}), so the failure is not
that the hierarchy is too coarse; hand-designing \emph{which} attribute should
win, however finely it is conditioned, is itself the wrong target.
Judge-labeled preference, not a hand-authored ranking, is the right target.

Aligning to that target nonetheless helps AHP as a side effect
(\Cref{fig:size_ahp}): with no AHP supervision at all, \model climbs from far
below chance to just above it at 8B, while the bases stay flat across sizes.
Judge-labeled preference training therefore recovers part of the intended priority
through scale---but only part; even the best model barely clears a coin flip,
so attribute hierarchy remains an open problem.

\subsection{Explicit budget: full analysis}
\label{app:budget}

Price was the worst attribute above, but the failure is one of supervision, not
capacity. Soft price preference is ambiguous---a higher price can signal
quality---so the judges that anchor our labels cannot give a clean target. A
\emph{stated} budget removes the ambiguity:
``\(\mathrm{price}\le\mathrm{budget}\)'' is a rule we can check automatically,
so we fine-tune the 0.6B base with the pairwise loss of \Cref{eq:dpo_reranker}
on budget-explicit pairs.

To test whether a model respects the budget rather than exploiting a shortcut,
the track has two slices (markers in \Cref{fig:diag_budget}). In the
\emph{threshold} slice, one product is under budget and the other over it, so
the under-budget product wins; in the \emph{control} slice, both are under
budget and a secondary attribute makes the \emph{pricier} product the winner.
The design is adversarial: always picking the cheaper product scores 100\% on
threshold but 0\% on control, and picking the pricier does the reverse---so
only genuine threshold behavior scores well on both.

The weakest open rerankers lean hard toward the pricier product (low threshold,
high control) and effectively ignore the cap; the Qwen bases sit between that
and threshold behaviour ($75.4/83.5$ at 8B, $67.4/82.6$ at 4B).
General preference alignment (\model) flips this---threshold accuracy
jumps---but overshoots toward cheap, so control dips below the bases: a soft
prefer-cheaper habit, not true threshold logic. This overshoot, rather than any
loss of capacity, is also why budget accuracy is not monotone in size:
\model-8B ($85.0\%$ overall) trails \model-4B ($88.6\%$) despite being the
better model on the threshold slice ($93.0$ vs.\ $92.6$), because its control
slice falls much further ($69.9$ vs.\ $80.8$). Scale sharpens the
prefer-cheaper lean; it does not install the threshold rule. Only the 0.6B
\emph{budget specialist}, trained on the automatic rule, is strong on both
slices and
highest overall (\(94.7\%\)). It is trained with general-relevance replay
mixed in rather than on budget pairs alone; we do not evaluate it on the
general benchmarks of \Cref{tab:mteb}. The conclusion is explicit: general preference alignment does not
install budget compliance, it only shifts the price bias, whereas a little
targeted, programmatically-labeled supervision recovers it almost entirely.
On this constraint, such targeted supervision beats both scale and general
alignment; we have not tested whether that transfers to other checkable
constraints, and the fully general case may still be best handled by a hard
budget filter at inference.

\subsection{Zero-shot AHP breakdown by hierarchy level}
\label{app:ahp_breakdown}

\Cref{tab:ahp_results} breaks the AHP diagnostic down by hierarchy level for
every system in \Cref{fig:diag_ahp}, computed on the same pairs with the same
harness.

The per-level view qualifies the aggregate result of \Cref{app:ahp}. Zero-shot,
\texttt{price} is not merely the hardest level but a near-total failure: every
open reranker scores at or near $0.000$, and the Qwen3-Reranker bases manage only
$0.000$, $0.177$ and $0.280$ at 0.6B, 4B and 8B, so all of them pick the
budget-violating product on a clear majority of pairs. Preference alignment fixes
precisely that---\model-4B reaches $0.973$ on price and $0.780$ on
color---but it simultaneously destroys \texttt{style} adherence, which falls
from $0.280$ to $0.015$ at 0.6B and from $0.200$ to $0.030$ at 4B, and costs
some \texttt{product type} accuracy. The aggregate climb toward chance is
therefore a redistribution across attributes rather than a uniform improvement,
which is further reason to read AHP as a diagnostic and not a target.

\begin{table}[h]
\centering
\small
\setlength{\tabcolsep}{4pt}
\begin{sc}
\begin{tabular}{lcccccc}
\toprule
& & \multicolumn{5}{c}{Accuracy by hierarchy level} \\
\cmidrule(lr){3-7}
Model & Overall & audience & prod.\ type & price & style & color \\
\midrule
\model-8B & \cellcolor{green!10}0.543 & 0.458 & 0.595 & 0.917 & 0.100 & 0.530 \\
\model-4B & \cellcolor{red!10}0.475 & 0.286 & 0.485 & \textbf{0.973} & 0.030 & \textbf{0.780} \\
\model-0.6B & \cellcolor{red!10}0.317 & 0.112 & 0.480 & 0.550 & 0.015 & 0.600 \\
\midrule
Jina-m0 (2.4B) & \cellcolor{red!10}0.299 & 0.286 & 0.635 & 0.003 & 0.215 & 0.070 \\
Qwen3-Rnk-4B & \cellcolor{red!10}0.278 & 0.148 & 0.620 & 0.177 & 0.200 & 0.020 \\
Qwen3-Rnk-8B & \cellcolor{red!10}0.275 & 0.118 & 0.598 & 0.280 & 0.110 & 0.090 \\
BM25 & \cellcolor{red!10}0.267 & 0.062 & \textbf{0.733} & 0.000 & 0.375 & 0.010 \\
Qwen3-Rnk-0.6B & \cellcolor{red!10}0.254 & 0.168 & 0.600 & 0.000 & 0.280 & 0.010 \\
BGE-v2-m3 (0.6B) & \cellcolor{red!10}0.244 & 0.224 & 0.552 & 0.000 & 0.165 & 0.000 \\
BGE-large (0.6B) & \cellcolor{red!10}0.233 & 0.208 & 0.495 & 0.000 & 0.230 & 0.010 \\
\bottomrule
\end{tabular}
\end{sc}
\caption{AHP diagnostic by hierarchy level (\(n=1{,}500\)), under the same
evaluation harness as \Cref{fig:diag_ahp}. Shading in the \textsc{overall} column marks
accuracy below the 0.50 chance line, i.e.\ systematic violation of the proposed
hierarchy; only \model-8B clears it (green). Per-level cells are unshaded, but
almost all of them are below chance too. Preference alignment does not lift the hierarchy uniformly---it
transforms \texttt{price} and \texttt{color} while \emph{degrading}
\texttt{style} and \texttt{product type}---which is why we treat the aggregate
AHP gain as a side effect rather than evidence of hierarchy-following.}
\label{tab:ahp_results}
\end{table}

\subsection{Full \texorpdfstring{\bench}{ShopRank-Bench} results}
\label{app:full_results}

\Cref{tab:main_structured,tab:main_nl} give the complete numbers behind
\Cref{fig:main_results}, and \Cref{tab:mcnemar} the paired tests behind every
significance claim in the paper.

\paragraph{Clustered and multiplicity-corrected tests.}
\phantomsection\label{app:significance}
Most queries contribute more than one pair---3.5 on average---and those pairs
share a candidate set and a single information need, so a bootstrap over
individual pairs would treat them as independent and report intervals that are
too narrow; we resample whole query groups instead. The paired McNemar tests of
\Cref{tab:mcnemar} make the opposite assumption, so we recomputed all eight
comparisons in both formats under a query-clustered bootstrap (20,000 resamples)
with a Holm--Bonferroni correction. The design effects are \(1.2\)--\(1.4\) and
no conclusion changes. The narrowest surviving margin is \model-4B over Jina-m0
on prose (\(p=9.0{\times}10^{-3}\)), and the \model-0.6B vs.\ base-4B
non-significance is likewise unaffected (\(p=0.11\) clustered, versus \(0.066\)
pair-level); we detect no difference but do not test equivalence against a
margin. That non-significance is specific to structured text, since on prose
base-4B leads by \(2.65\) points (\(p<10^{-4}\)). The raw and Holm-adjusted
\(p\)-values and the design effects are produced and recorded by the
released evaluation code. Intervals elsewhere
are query-clustered bootstrap intervals unless identified otherwise;
\Cref{app:fourth_judge} reports a Wilson interval over an independent 45-pair
sample.

\begin{table}[h]
\centering
\small
\setlength{\tabcolsep}{4pt}
\begin{sc}
\begin{tabular}{lcccc}
\toprule
Model & Overall [95\% CI] & Gold & \makecell{Silver\\(2/3)} & \makecell{Bronze\\(1/3)} \\
\midrule
\model-8B & \cellcolor{green!10}\textbf{83.4 [82.5, 84.2]} & \textbf{96.5} & \textbf{86.1} & \textbf{74.8} \\
\model-4B & \cellcolor{green!10}81.2 [80.3, 82.0] & 96.3 & 84.3 & 71.4 \\
Qwen3-Rnk-8B & 79.2 [78.2, 80.1] & 93.2 & 82.6 & 69.4 \\
Jina-m0 (2.4B) & 79.2 [78.3, 80.1] & 90.8 & 82.9 & 70.2 \\
Qwen3-Rnk-4B & 77.5 [76.5, 78.5] & 92.7 & 81.4 & 66.7 \\
\model-0.6B & 76.6 [75.7, 77.6] & 91.2 & 79.7 & 67.1 \\
Qwen3-Rnk-0.6B & 73.5 [72.4, 74.5] & 85.7 & 77.6 & 63.7 \\
BGE-v2-m3 (0.6B) & 73.3 [72.3, 74.3] & 84.9 & 77.8 & 63.5 \\
BGE-large (0.6B) & 71.6 [70.5, 72.7] & 82.4 & 76.0 & 62.2 \\
BM25 & 56.2 [54.9, 57.4] & 69.0 & 60.4 & 46.2 \\
\midrule
\multicolumn{5}{l}{\emph{Reference: zero-shot reasoning LLMs, not rerankers (\Cref{app:llm_reference})}} \\
GLM-5.3 & 91.9 [91.2, 92.4] & 99.6 & 95.9 & 84.3 \\
Qwen3.5-27B & 92.1 [91.5, 92.7] & 99.9 & 96.0 & 84.6 \\
\bottomrule
\end{tabular}
\end{sc}
\caption{\bench structured format (\(n=10{,}511\)). Accuracy is in percent;
intervals are 95\% query-clustered bootstrap CIs. Tiers are by the number of
judge families that committed (\Cref{sec:benchmark_construction}).
\textsc{Qwen3-Rnk} rows are the un-fine-tuned Qwen3-Reranker bases evaluated
zero-shot under \Cref{lst:qwen_prompt}; each \model{} row is the LoRA-aligned
counterpart of the base of the same size. Every reranker loses 20--26 points
from gold to bronze, which is why the aggregate column alone is a poor summary.
Bold marks the best reranker; the LLM rows count a pair correct only when both
presentation orders pick the preferred product.}
\label{tab:main_structured}
\end{table}

\begin{table}[h]
\centering
\small
\setlength{\tabcolsep}{4pt}
\begin{sc}
\begin{tabular}{lcccc}
\toprule
Model & Overall [95\% CI] & Gold & \makecell{Silver\\(2/3)} & \makecell{Bronze\\(1/3)} \\
\midrule
\model-8B & \cellcolor{green!10}\textbf{81.8 [80.9, 82.7]} & 95.0 & \textbf{85.4} & \textbf{72.2} \\
\model-4B & \cellcolor{green!10}79.5 [78.6, 80.4] & \textbf{95.4} & 82.1 & 69.8 \\
Qwen3-Rnk-8B & 78.4 [77.5, 79.4] & 91.9 & 82.3 & 68.4 \\
Jina-m0 (2.4B) & 78.1 [77.1, 79.0] & 88.8 & 82.7 & 68.5 \\
Qwen3-Rnk-4B & 77.9 [76.9, 78.9] & 92.1 & 81.0 & 68.5 \\
\model-0.6B & 75.3 [74.2, 76.3] & 91.2 & 78.1 & 65.4 \\
BGE-v2-m3 (0.6B) & 73.4 [72.4, 74.5] & 85.8 & 77.4 & 63.9 \\
BGE-large (0.6B) & 73.0 [71.9, 74.0] & 85.6 & 77.3 & 63.0 \\
Qwen3-Rnk-0.6B & 72.6 [71.6, 73.7] & 84.5 & 77.0 & 62.9 \\
BM25 & 58.4 [57.1, 59.6] & 70.8 & 61.4 & 49.7 \\
\bottomrule
\end{tabular}
\end{sc}
\caption{Natural-language rendering of the same \bench pairs, under the same
tiers. Note the ordering change against \Cref{tab:main_structured}: base-8B
overtakes Jina-m0 on prose, while base-0.6B falls below both BGE baselines.
BM25 is below chance on the bronze tier in both formats.}
\label{tab:main_nl}
\end{table}

\begin{table}[h]
\centering
\small
\setlength{\tabcolsep}{5pt}
\begin{sc}
\begin{tabular}{llc}
\toprule
Track (format) & Comparison & Paired test \\
\midrule
Preference (structured) & 4B vs.\ Jina-m0 & \(\chi^2=18.2,\ p=1.9{\times}10^{-5}\) \\
Preference (structured) & 8B vs.\ Jina-m0 & \(\chi^2=85.0,\ p=3.0{\times}10^{-20}\) \\
Preference (structured) & 8B vs.\ 4B & \(\chi^2=29.7,\ p=5.0{\times}10^{-8}\) \\
Preference (structured) & 0.6B vs.\ BGE-v2-m3 & \(\chi^2=47.4,\ p=5.8{\times}10^{-12}\) \\
Preference (structured) & 0.6B vs.\ base-0.6B & \(\chi^2=68.9,\ p=1.0{\times}10^{-16}\) \\
Preference (structured) & 4B vs.\ base-4B & \(\chi^2=84.0,\ p=4.8{\times}10^{-20}\) \\
Preference (structured) & 8B vs.\ base-8B & \(\chi^2=135.3,\ p=2.8{\times}10^{-31}\) \\
Preference (structured) & 0.6B vs.\ base-4B & \(\chi^2=3.4,\ p=0.066\) (n.s.) \\
\midrule
Preference (NL) & 4B vs.\ Jina-m0 & \(\chi^2=8.6,\ p=3.3{\times}10^{-3}\) \\
Preference (NL) & 8B vs.\ Jina-m0 & \(\chi^2=68.7,\ p=1.1{\times}10^{-16}\) \\
Preference (NL) & 0.6B vs.\ BGE-v2-m3 & \(\chi^2=14.3,\ p=1.6{\times}10^{-4}\) \\
Preference (NL) & 0.6B vs.\ base-0.6B & \(\chi^2=41.7,\ p=1.1{\times}10^{-10}\) \\
Preference (NL) & 4B vs.\ base-4B & \(\chi^2=14.6,\ p=1.3{\times}10^{-4}\) \\
Preference (NL) & 8B vs.\ base-8B & \(\chi^2=113.0,\ p=2.2{\times}10^{-26}\) \\
Preference (NL) & 0.6B vs.\ base-4B & \(\chi^2=33.1,\ p=8.8{\times}10^{-9}\)\textsuperscript{\dag} \\
\midrule
AHP diagnostic & 0.6B vs.\ base-0.6B & \(\chi^2=17.9,\ p=2.4{\times}10^{-5}\) \\
AHP diagnostic & 4B vs.\ base-4B & \(\chi^2=135.1,\ p=3.1{\times}10^{-31}\) \\
AHP diagnostic & 8B vs.\ base-8B & \(\chi^2=295.6,\ p=3.0{\times}10^{-66}\) \\
\midrule
Budget diagnostic & 0.6B vs.\ base-0.6B & \(\chi^2=382.0,\ p=4.6{\times}10^{-85}\) \\
Budget diagnostic & 4B vs.\ base-4B & \(\chi^2=150.0,\ p=1.8{\times}10^{-34}\) \\
Budget diagnostic & 8B vs.\ base-8B & \(\chi^2=34.4,\ p=4.4{\times}10^{-9}\) \\
\bottomrule
\end{tabular}
\end{sc}
\caption{Continuity-corrected paired McNemar tests across all \bench tracks.
Unqualified model names denote \model variants; ``base'' denotes the
corresponding un-fine-tuned Qwen3-Reranker.
\textsuperscript{\dag}Significant in favor of base-4B: the absence of a
detected 0.6B/4B difference is specific to the structured format.}
\label{tab:mcnemar}
\end{table}

\subsection{Further limitations}
\label{app:limitations}

Beyond the label-provenance limitation of \Cref{sec:conclusion}: the judge panel
is not uniformly active, leaving the single-judge bronze tier largely
Gemma-decided (\Cref{sec:benchmark_construction}); judges return a unanimous tie or an
outright conflict on $54.3\%$ of hard adjacent pairs ($53.7\%$ and $0.6\%$
respectively), capping the supply of clean on-policy labels; a gap to zero-shot reasoning LLMs remains open
(\Cref{app:llm_reference}); mixed-format training, added for robustness to
serialization, is neutral on \bench{} (\Cref{app:format_training}); the
natural-language view is LLM-rendered rather than scraped prose, and varies
attribute inclusion as well as phrasing (\Cref{sec:dual_format});
public-benchmark comparisons may reflect pretraining contamination
(\Cref{tab:mteb}); the category mix leans toward apparel
(\Cref{app:categories}); and the queries are English- and US-centric.

\end{document}